\documentclass[10pt,twocolumn,letterpaper]{article}

\usepackage[pagenumbers]{cvpr} %
\usepackage{xspace}
\def\Ours{GeoDream\xspace}
\usepackage[table]{xcolor}%
\usepackage[ruled,linesnumbered]{algorithm2e} %
\usepackage{float}
\usepackage{placeins}

\definecolor{cvprblue}{rgb}{0.21,0.49,0.74}
\usepackage[pagebackref,breaklinks,colorlinks,citecolor=cvprblue]{hyperref}

\title{\Ours: Disentangling 2D and Geometric Priors for High-Fidelity and Consistent 3D Generation}

\author{
Baorui Ma$^{1}$\thanks{Equal contribution.\\
Correspondence to \textit{\{brma@baai.ac.cn\}} and \textit{\{wangxinlong@baai.ac.cn.\}}}  ~
Haoge Deng \textsuperscript{2,1}$^*$~
Junsheng Zhou$^{3, 1}$~
Yu-Shen Liu$^{3}$ ~
Tiejun Huang$^{1,4}$ ~
Xinlong Wang$^{1}$
\\
$^1$ Beijing Academy of Artificial Intelligence \quad
$^2$ BUPT \quad
$^3$ Tsinghua University \quad
$^4$ Peking University
}

\begin{document}
\maketitle
\begin{abstract}
Text-to-3D generation by distilling pretrained large-scale text-to-image diffusion models has shown great promise but still suffers from inconsistent 3D geometric structures (Janus problems) and severe artifacts. The aforementioned problems mainly stem from 2D diffusion models lacking 3D awareness during the lifting. In this work, we present \Ours, a novel method that incorporates explicit generalized 3D priors with 2D diffusion priors to enhance the capability of obtaining unambiguous 3D consistent geometric structures without sacrificing diversity or fidelity. Specifically, we first utilize a multi-view diffusion model to generate posed images and then construct cost volume from the predicted image, which serves as native \textbf{3D geometric priors}, ensuring spatial consistency in 3D space. Subsequently, we further propose to harness 3D geometric priors to unlock the great potential of 3D awareness in 2D diffusion priors via a disentangled design. Notably, disentangling 2D and 3D priors allows us to refine 3D geometric priors further. We justify that the refined 3D geometric priors aid in the 3D-aware capability of 2D diffusion priors, which in turn provides superior guidance for the refinement of 3D geometric priors. Our numerical and visual comparisons demonstrate that \Ours generates more 3D consistent textured meshes with high-resolution realistic renderings (i.e., 1024 $\times$ 1024) and adheres more closely to semantic coherence. Our code and evaluation of 3D metric are available at: \href{https://mabaorui.github.io/GeoDream_page}{GeoDream}

\end{abstract}    

\section{Introduction}
\label{sec:intro}

Diffusion models~\cite{saharia2022photorealistic,rombach2022high,ramesh2022hierarchical} have significantly advanced text-to-image synthesis. This remarkable achievement has been reached by training scalable generative models on a vast corpus of paired text-image data. Inspired by their success, it is appealing to lift this success from 2D to 3D because this achievement holds significant potential impacts on the modern game and media industry. Template-based generators~\cite{chen2023fantasia3d} and 3D native generative models~\cite{li2023diffusion,yu2023pushing,mo2023dit,nichol2022point,jun2023shap} provide a natural and direct approach to the lift.  
However, due to the massive and diverse 3D data required to train such generalized models, these methods are usually limited to specific categories with relatively simple topology and texture. Recently, the Score Distillation Sampling (SDS)~\cite{poole2022dreamfusion} and Variational Score Distillation (VSD)~\cite{wang2023prolificdreamer} 
have been introduced to optimize 3D representations such that images rendered from any viewpoints maintain a high likelihood, as evaluated by diffusion model conditioned on a given text. This is an exciting direction because it allows for generating 3D assets from any given text prompts, circumventing the need for any 3D data. Despite these methods yielding satisfactory results on a wide range of geometrically symmetrical 3D shapes, empirical observations indicate that SDS and VSD losses still suffer from inconsistent 3D geometric structures (Janus problems)~\cite{wiki:janus} and severe artifacts~\cite{wang2023prolificdreamer,shi2023mvdream} with asymmetric geometry. This is primarily due to the lack of 3D awareness in 2D diffusion models, which inherently makes the lifting from 2D observations into 3D ambiguous.

\begin{figure*}[htb]
    \centering
    \includegraphics[width=0.975\textwidth]{fig/Fig1.pdf}
    \caption{ \Ours alleviates the Janus problems by incorporating explicit 3D priors with 2D diffusion priors. \Ours generates consistent multi-view rendered images and rich details textured meshes. We remove rendering background to achieve a clearer visualization.}
    \label{fig:fig1}
    \vspace{-0.4cm}
\end{figure*}

As a remedy, learning 3D priors from 3D datasets seems theoretically reasonable and correct. However, 3D data remains expensive and sparse compared to the plentifully available images. Therefore, the most promising avenue~\cite{qian2023magic123,shi2023mvdream,sun2023dreamcraft3d} presently is to equip 2D diffusion priors with 3D priors learned from relatively limited 3D data, aiming to achieve the best of both worlds. Recently, with the release of large-scale 3D datasets, Objaverse~\cite{deitke2023objaverse} and Objaverse-XL~\cite{deitke2023objaverse-xl}, a few works~\cite{liu2023zero,li2023sweetdreamer,shi2023mvdream,ye2023consistent} have attempted to finetune pre-trained 2D diffusion models using multi-view images rendered from 3D dataset. This involves obtaining multi-view images from the fine-tuned diffusion model conditioned on camera parameters and utilizing the clues of predicted multi-view consistency to infer 3D information. Nevertheless, these methods rely heavily on the consistency of content predicted across different 
 source views. Despite their efforts to employ 3D self-attention to exchange features between different views~\cite{shi2023mvdream}, to correlate multi-view features using 3D-aware attention~\cite{ye2023consistent}, or to transform RGB predictions into coarser Canonical Coordinates Map predictions~\cite{li2023sweetdreamer} to mitigate the negative impact of inconsistencies. Such inconsistencies between the predicted multiple views become particularly noticeable, especially in imaginative and uncommon cases beyond the training data distribution, resulting in over-smoothing and the loss of semantic geometries in the generated 3D assets.

To resolve this issue, we introduce \Ours, a novel method that incorporates explicit generalized 3D priors with 2D diffusion priors to enhance the capability of obtaining unambiguous 3D consistent geometric structures, while maintaining diversity and high fidelity. Our contributions are listed below. i) In stark contrast to the methods mentioned above that hinge heavily upon the consistency between multi-view priors, we propose to obtain 3D native priors within the 3D world space, which are well-suited to handle the inherent lack of perfect consistency within the multi-view predicted priors, and naturally free from inconsistencies caused by camera viewpoint transition. ii) We justify that disentangling 3D and 2D priors is a potentially exciting direction for maintaining both the generalization of 2D diffusion priors and the consistency of 3D priors. In other words, providing hints through 3D priors to unlock the great potential of 3D awareness in 2D diffusion priors, without the need for invasive finetune 2D diffusion models.

Specifically, we start by reconstructing cost volume as native 3D priors by aggregating the predicted multi-view 2D images into 3D space. Such aggregation operations have been widely used in MVS-based techniques~\cite{yao2018mvsnet,zhang2022nerfusion,long2022sparseneus,liu2023one}, which are known to be robust and generalized to provide valuable cues for geometric reasoning. We find that such operations are well-suited for handling imperfect and inconsistent multi-view predictions. The reason is that they involve multi-view information aggregation, which helps filter out inconsistent content to some extent, rather than dealing with each view individually. Foremost, we conduct extensive experiments to demonstrate that our proposed 3D priors adapt to multiple views predicted by various off-the-shelf multi-view diffusion models, such as Zero123~\cite{liu2023zero}, MVDream~\cite{shi2023mvdream} and Zero123++~\cite{shi2023zero123++}. Moreover, we introduce a critical viewpoint sampling strategy to promote the stability of the 3D priors.

We further propose incorporating 3D priors with 2D diffusion priors in a disentangled solution. Existing multi-view diffusion priors are equipped with 2D diffusion priors in a coupled way, including generating multiple views as supervision~\cite{liu2023zero,shi2023zero123++} or distilling the probability density as a loss~\cite{shi2023mvdream,li2023sweetdreamer,sun2023dreamcraft3d,qian2023magic123} to compute gradients for optimizing 3D representations. Instead, we justify that leveraging the geometric clues provided by 3D priors can effectively unleash the great potential 3D awareness capability inherent in 2D diffusion priors, referred to as  ``disentangled design". 
Very recent works have started to explore how to evoke 3D-aware ability in 2D diffusion by altering score functions~\cite{hong2023debiasing} or negative text prompts~\cite{armandpour2023re}. These efforts have made surprising progress, yet the performance remains unstable regarding 3D consistency. 
Our insight is that going through geometric priors to unlock the great potential of 3D awareness in 2D diffusion is a promising direction that is both general and stable. 
Moreover, we rely solely on the awakened 3D-aware capability of 2D priors to guide the optimization of Neural Implicit Surfaces (NeuS)~\cite{wang2021neus} without the supervision of 3D priors, thereby avoiding compromising the inherent advantages of 2D priors in terms of generalization and creativity. 
We show that 3D priors can be further refined to boost rendering quality and geometric accuracy. The 2D diffusion priors benefit from gradually evolved 3D priors, which in turn provide superior guidance for unleashing the 2D priors. Finally, we use DMTet~\cite{shen2021deep} to extract textured mesh from optimized NeuS for mesh fine-tuning. 
Unlike previous work~\cite{poole2022dreamfusion,wang2023prolificdreamer} attempt to increase the rendering resolution, which typically suffer from over-saturation issues, we successfully increase the rendering resolution from 512 to 1024.
We hypothesize that the improved results are aided and abetted by 3D priors that provide more plausible geometry and realistic texture, making the optimization easier, because the rendered image is closer to diffused distributions. 
To comprehensively evaluate semantic coherence, to our knowledge, we are the first to propose $\rm{Uni3D_{score}}$ metric, lifting the measurement from 2D to 3D.

\begin{table}
  \caption{Comparison of design space.}
  \vspace{-0.35cm}
  \label{tab:Design_space}
  \centering
  \scalebox{0.8}{
  \setlength{\tabcolsep}{0mm}{
  \begin{tabular}{l c c c c }
    \hline
    \bf Method & \bf Prolific~\cite{wang2023prolificdreamer} & \bf MVDream~\cite{shi2023mvdream} & \bf GSGEN~\cite{chen2023text} & \bf Ours \\
    \midrule
    Repr. & NeRF+DMTet & NeRF & Gaussian & NeuS+DMTet \\
     Resolution  & 512 & 512 & 512 & 1024 \\
    3D guidance  & - & Multi-Views & Point-E & Cost volume \\
    3D\&2D & - & Entangled & Entangled & Disentangled \\
    3D priors & Fixed & Fixed & Fixed & Optimizable \\
    \hline
  \end{tabular}
  }
  }
  \vspace{-0.65cm}
\end{table}

As summarized in Tab.\ref{tab:Design_space}, we compared the latest methods\cite{wang2023prolificdreamer,shi2023mvdream,chen2023text} in design space, including 3D representation, rendering resolution, forms of 3D guidance, the disentangling of 3D and 2D priors and the optimizability of 3D priors. As shown in Fig.\ref{fig:fig1}, \Ours can yield 1024 $\times$ 1024 high-resolution rendered images and high-fidelity textured meshes while greatly alleviating the notorious Janus problems. In Sec.\ref{sec.result}, we conduct comprehensive evaluations that demonstrate the superiority of the 3D assets generated by \Ours in terms of plausible geometry and delicate rendering details in visual appearance.

\section{Related Work}

\noindent\textbf{3D Generation Guided by 2D Priors.}
Deep generative models have driven the field of 3D generation. Some efforts utilize Variational Auto Encoders (VAEs)~\cite{kingma2013auto} for texture generation~\cite{henderson2020learning,henderson2020leveraging}, while Generative Adversarial (GAN) Models~\cite{goodfellow2014generative} investigate 3D-aware GAN training~\cite{chan2022efficient,deng2022gram}. Thus far, diffusion models have exhibited relatively better generalizability and training stability for diverse object generation compared to GANs and VAEs, and thus have gradually become recent focal points in 3D generation. Specifically, recent endeavors attempt to leverage the potent 2D diffusion priors to aid 3D generation by coupling it with a 3D representation, such as NeRF~\cite{mildenhall2021nerf}, DMTNet~\cite{mildenhall2021nerf}, or NeuS~\cite{wang2021neus}, among others, bypasses the necessity for extensive text-3D datasets for training 3D generative models. Such methods involve various techniques, including score distillation sampling schedules like SDS~\cite{wang2023score}, SJC~\cite{poole2022dreamfusion}, and VSD~\cite{wang2023prolificdreamer} losses, which optimize the 3D representation by enhancing high likelihood evaluated by the 2D diffusion models. A coarse-to-fine training strategy~\cite{chen2023fantasia3d} strengthens texture representation, decoupling geometric and texture aspects of 3D representation for finer optimization~\cite{lin2023magic3d}, improving 3D representation~\cite{tang2023dreamgaussian,chen2023text}. Although these methods demonstrate the ability to generate photo-realistic and diverse 3D assets with user-provided textual prompts, they are prone to the notorious 3D inconsistency issues (Janus problems) during the lifting process due to their reliance on 2D diffusion models for training, which lack 3D knowledge. Despite some current methods attempting to address 3D inconsistency by altering score functions~\cite{hong2023debiasing} or negative text prompts~\cite{armandpour2023re}, performance remains instability in terms of 3D consistency. In this work, we aim to explore the distinctive advantages of incorporating explicit 3D priors with 2D priors, enabling the generation of highly detailed 3D objects while remarkably mitigating 3D inconsistency issues.

\noindent\textbf{3D Generation Guided by 3D Priors.}
Learning 3D priors from 3D datasets seems theoretically reasonable and correct for enhancing the coherency of 3D generation~\cite{liu2023zero,liu2023one,lin2023magic3d,melas2023realfusion,xu2023neurallift,purushwalkam2023conrad}. Therefore, various 3D latent diffusion models trained on 3D data have been recently introduced, including those using Tri-plane~\cite{shue20233d} or feature grid~\cite{wang2023rodin,karnewar2023holodiffusion,liu2023syncdreamer} encoding 3D representations into the latent space. Additionally, OpenAI has explored models aiming to directly generate 3D formats using several million internal 3D shapes, such as point clouds~\cite{nichol2022point} or neural radiance fields~\cite{jun2023shap}. However, their generalizability to the scope of their 2D counterparts remains unverified, due to the relative sparsity of 3D data compared to the abundance of available 2D images. 
Consequently, the most promising avenue currently is to equip 2D diffusion priors with 3D priors learned from relatively limited 3D data, intending to achieve the best of both worlds. Recently, with the release of a large-scale 3D dataset called Objaverse~\cite{deitke2023objaverse} and Objaverse-XL~\cite{deitke2023objaverse-xl}, some work~\cite{liu2023zero,yang2023consistnet,liu2023one,li2023sweetdreamer,shi2023mvdream,ye2023consistent} has attempted to fine-tune pre-trained 2D diffusion models using multi-view images rendered from 3D data. This aims to generate multi-view images from the fine-tuned diffusion model conditioned on camera parameters and utilize the clues of predicted multi-view consistency to assist in inferring 3D information. Nevertheless, these methods heavily depend on the absolute consistency of content predicted across different views. Nonetheless, their efforts to utilize 3D self-attention~\cite{shi2023mvdream,yang2023consistnet} for feature exchange between different views, to correlate multi-view features using 3D-aware attention~\cite{ye2023consistent}, or to transform RGB predictions into coarser Canonical Coordinates Map predictions~\cite{li2023sweetdreamer} for mitigate the negative impact of inconsistencies. The performance of such methods frequently exacerbates inconsistencies and unrealistic rendering quality in uncommon cases, due to the absence of explicit constraints between different predicted viewpoints within 3D space. In this work, we incorporate explicit generalized 3D priors into 2D diffusion priors. These explicit 3D priors fundamentally ensure consistency in 3D space and avoid the independence of multi-view priors across source views. 

\begin{figure}[tb]
    \centering
    \includegraphics[width=\linewidth]{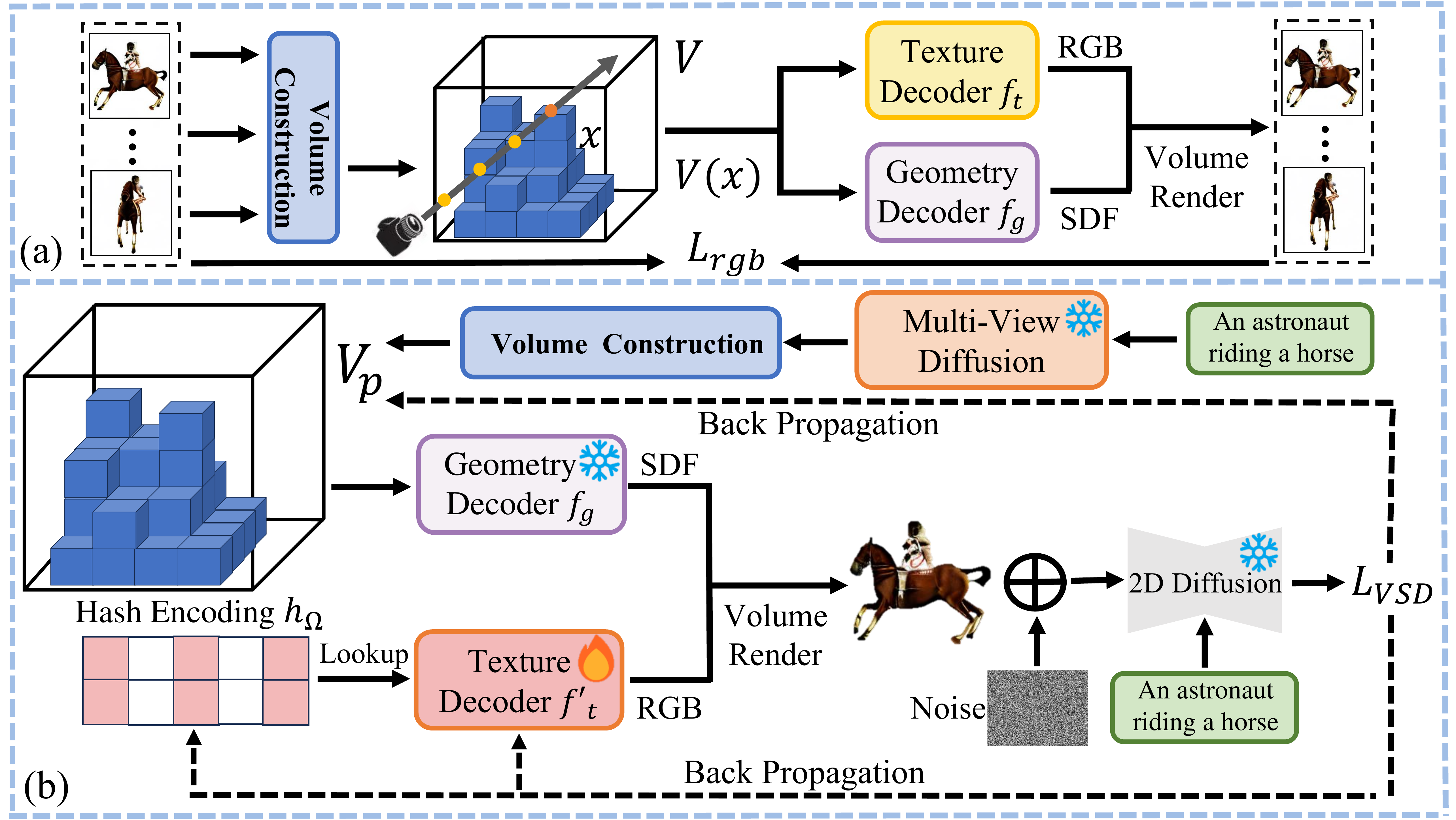}
    \vspace{-0.7cm}
    \caption{The overview of \Ours. (a) 3D priors training. (b) Incorporating 3D priors with 2D diffusion priors.}
    \label{fig:overview}
    \vspace{-0.65cm}
\end{figure}

\section{Method}
\label{sec.method}

We focus on generating 3D content with consistently accurate geometry and delicate visual detail, by equipping 2D diffusion priors with the capability to produce 3D consistent geometry while retaining their generalizability. The overview
of \Ours is shown in Fig.\ref{fig:overview}. \Ours consists of the following two stages. i) During 3D priors training, we build upon the One-2-3-45~\cite{liu2023one}, which encodes geometry into cost volume $V$ and geometry MLP decoder $f_{g}$. In addition, the appearance of the object is modeled to cost volume $V$ and texture MLP decoder $f_{t}$. We refer to the trained geometric decoder $f_{g}$ and appearance decoder $f_{t}$ with cost volume $V$ as native 3D geometric priors and appearance priors, as shown in Fig.\ref{fig:overview} (a). Detials in Sec.\ref{sec.stage1}.  ii) During priors refinement, we show that geometric priors can be further fine-tuned to boost rendering quality and geometric accuracy by combining a 2D diffusion model, as shown in Fig.\ref{fig:overview} (b). Detials in Sec.\ref{sec.stage2}.

\subsection{Generalizable 3D Priors Training}
\label{sec.stage1}
We start by reconstructing cost volume $V$ as native 3D priors by aggregating the 2D image features into 3D space, which provides valuable cues for geometric reasoning in the priors refinement stage.

\noindent\textbf{Cost Volume Construction.} 
Following MVS-based methods~\cite{yao2018mvsnet,zhang2022nerfusion,long2022sparseneus,liu2023one}, given multi-view images $I=\{(I_{i})^{N-1}_{i=0}\}$, we extract 2D feature maps $F=\{(F_{i})^{N-1}_{i=0}\}$ using a 2D feature network $f_{2D}$. Volume reconstruction model takes posed 2D feature maps $F$ as input and outputs cost volume $V$ with per-voxel neural features in voxels. Specifically, for each voxel centered at 3D location $h$, per-voxel neural feature is computed by projecting each location $h$ to $N$ image feature planes and then fetching the variance of the features at the the location of the projection. We use $\mathrm{~Var}$ to denote the variance operation and $P$ to denote the  projection procedure. We then use a sparse 3D CNN $f_{3D}$ to process the variance features per voxel to regress the cost volume, as formulated by, 
\begin{equation}
\label{eq:cost}
V = f_{3D}(\mathrm{~Var}\{P(F_{i},  h)\}_{i=0}^{N-1}),
\end{equation}

\noindent where the variance operation is invariant to the number $N$ of input images. We find that such an operation is well-suited for handling imperfect and inconsistent multi-view predictions, due to involving information aggregation rather than dealing with each view individually.

\noindent\textbf{Geometry and Texture Decoder.}
The cost volume $V$ is directly decoded into signed distance function values (SDF) and color information using the corresponding geometry MLP decoder $f_{g}$ and texture MLP decoder $f_{t}$. For any arbitrary query point $x\in \mathbb{R}^3$, we get the SDF $s$ and color $c$ as
\begin{equation} 
\label{eq:sdf}
s(x)\,=\,f_{g}(E(x), V(x)),
\end{equation} 
\begin{equation}
\label{eq:color-pre}
c(x)=f_{t}(\{P(F_{i},  x)\}_{i=0}^{N-1},V(x),\{\Delta d_{i}\}_{i=0}^{N-1}),
\end{equation} 

\noindent where $E$ denotes position encoding, $V(p)$ denotes tri-linearly interpolated feature from cost volume at query point $x$, 
$\Delta d_{i} = d - d_{i}$ is the viewing direction of the query ray relative to the viewing direction of the $i{th}$ multi-view image.

The final rendered image $I{'}$ is achieved via SDF-based differentiable volume rendering $R$.  In this work, we get the pre-trained parameters of the $f_g$, $f_t$, and $f_{3D}$ networks from the One-2-3-45~\cite{liu2023one}, which is trained on ground truth images $I$ rendered from the Objaverse dataset with a loss
\begin{equation}
\mathcal{L}_{rgb} = ||I-I{'}||_{2}, 
\end{equation}
\noindent where $I{'}=R(\{s(x_j),c(x_j))\}_{j=0}^{M-1})$, $M$ denotes sampling $M$ query points along the ray of viewing direction.

\subsection{Priors Refinement}
\label{sec.stage2}
We present how to further finetune the geometric priors obtained from 3D priors training stage, i.e., optimizable cost volume $V$ and the fixed pre-trained geometric decoder $f_{g}$, using the 2D diffusion priors, as shown in Fig.\ref{fig:overview} (b). During priors refinement stage, we replace the $N$ ground truth rendered images with multi-view diffusion model predictions. In contrast to One-2-3-45, \Ours 
is not limited to the Zero123~\cite{liu2023zero} predictions. We conduct extensive experiments with various multi-view diffusion models, such as MVDream~\cite{shi2023mvdream} and Zero123++~\cite{shi2023zero123++}. We also introduce a critical viewpoint sampling strategy to ensure \Ours robustly adapts to various multi-view diffusion models, rather than being limited to just one. Overall, we justify that by decoupling 3D and 2D diffusion priors, \Ours unlocks the immense potential of 3D awareness in the 2D diffusion model, avoiding the tendency to produce canonical views, resulting in 3D assets featuring multiple faces and collapsed geometry. Thanks to the decoupling, \Ours maintains the generalization and imaginativeness of 2D diffusion priors, while also exploring the significant role that geometric priors play in improving appearance modeling.

\noindent\textbf{Multi-View Images Generation.} The rapid advancement of 3D generation has provided a wide range of methods available for generating multi-view images, such as Zero123\cite{liu2023zero}, MVDream\cite{shi2023mvdream}, and Zero123++~\cite{shi2023zero123++}. Given a set of predefined camera poses $\{(R_{i},T_{i})^{N-1}_{i=0}\}$ and a user-provided condition $c$, we utilize a fixed multi-view diffusion $f_{mv}$ to predict posed images $I_p=\{(I^p_{i})^{N-1}_{i=0}\}$ and  extract 2D feature maps $F_p=\{(F^p_{i})^{N-1}_{i=0}\}$, 
\begin{equation}
F^p_{i}=f_{2D}(f_{mv}(c,R_{i},T_{i})), 
\end{equation}
\noindent where $R\in\mathbb{R}^{3\times3}$, $T\in\mathbb{R}^{3\times3}$ respectively denote relative camera rotation and translation of the default viewpoint.

\noindent\textbf{3D Geometric Priors.}
By replacing $F_{i}$ in Eq.\ref{eq:cost} into $F^p_{i}$, we obtain the value of SDF at an arbitrary query point $x$ defined in Eq.\ref{eq:sdf},
\begin{equation}
\label{eq:cost-predict}
V_p = f_{3D}(\mathrm{~Var}\{P(F^p_{i},  h)\}_{i=0}^{N-1}),
\end{equation}
\begin{equation} 
\label{eq:sdf-prior}
s_p(x)\,=\,f_{g}(E(x), V_p(x)),
\end{equation} 
\noindent where $s_p(x)$ is treated as a geometric prior since it encodes the hidden geometric clues in the predicted multiple views.

\noindent\textbf{Texture Decoder.}
We propose to drop the pre-trained texture priors $f_t$ defined in Eq.\ref{eq:color-pre} because we empirically find that texture priors tend to generate 3D assets with lighting and texture styles similar to the rendered dataset. We choose Instant NGP~\cite{muller2022instant} for efficient high-resolution texture encoding. Specifically, for any arbitrary query point $x\in \mathbb{R}^3$, a learnable hash encoding $h_{\Omega}$ is decoded into a color $c$ using initialized texture decoder $f_{t}'$,  as formulated by, 
\begin{equation} 
\label{eq:color}
c_p(x)=
f_{t}'
(h_{\Omega}(x), x),
\end{equation}
\noindent where $h_{\Omega}(x)$ denotes the looked-up feature vector from $h_{\Omega}$ at query point $x$.

\noindent\textbf{Texture and Geometry Refinement.} 
To incorporate 3D geometric priors with 2D diffusion priors, we minimize the VSD loss introduced in ProlificDreamer~\cite{wang2023prolificdreamer} to optimize the parameters of ${\theta}_{1}$ in cost volume $V$, ${\theta}_{2}$ in hash encoding $h_{\Omega}$ and ${\theta}_{3}$ in texture decoder $f_{t}'$. At each iteration, we sample a camera pose $o$ from a pre-defined distribution. We render 2D image $\hat{x}$ at pose $o$ by combining Eq.\ref{eq:sdf-prior} and Eq.\ref{eq:color} via differential rendering $R$. Our objective function during priors refinement is to minimize the VSD loss $\mathcal{L}_{VSD}$, the corresponding gradient $\nabla_{{\theta}_{1},{\theta}_{2},{\theta}_{3}}\mathcal{L}_{VSD}$ is 
\begin{equation} 
\label{eq.loss}
\mathrm{E}_{t, \epsilon, o} [w(t)(\epsilon_{pretrain}(\hat{x_{t}}, t, c) - \epsilon_{l}(\hat{x_{t}}, t, c, o))\frac{\partial \hat{x}}{\partial ({\theta}_{1},{\theta}_{2},{\theta}_{3})}],
\end{equation} 
\noindent where $\hat{x_{t}}$ denotes a noisy rendered image in timestep $t$, $w(t)$ denotes a weighting function, $\epsilon_{pretrain}$ is a 2D pretrained diffusion model and $\epsilon_{l}$ is a trainable LoRA~\cite{hu2021lora} diffusion model with parameters of $l$.  We propose to fix the geometry decoder $f_g$ conjointly with a learning rate decay strategy for the cost volume, aiming to maintain geometric priori cues as well as tuning to achieve better details in the early stage of optimization. More details on viewpoint sampling and learning rate decay strategy are provided in Sec.\ref{sec.ablation}.

\noindent\textbf{Mesh Fine-tuning.}
For high-resolution rendering, we use DMTet~\cite{shen2021deep} to extract textured 3D mesh representation from optimized NeuS~\cite{wang2021neus}. By minimizing the loss in Eq.\ref{eq.loss}, we follow ProlificDreamer~\cite{wang2023prolificdreamer} first to optimize the geometry using the normal map and then optimize the texture. We empirically find that we can increase the rendering resolution from 512 to 1024. But unlike previous work~\cite{poole2022dreamfusion,wang2023prolificdreamer}, attempting to increase the rendering resolution suffers from over-saturation issues. 
We successfully increase the rendering resolution from 512 to 1024.
We hypothesize that well-optimized results are aided and abetted by 3D priors that provide more plausible geometry and realistic texture, making the optimization easier, because the rendered image $\hat{x}$ is closer to diffused distributions at each iteration.

\section{Experiment}
\subsection{Results of \Ours}
\label{sec.result}

\noindent\textbf{Baselines.} We report our performance with the latest 3D generation methods, including DreamFusion~\cite{poole2022dreamfusion}, ProlificDreamer~\cite{wang2023prolificdreamer}, MVDream~\cite{shi2023mvdream}, GSGEN~\cite{chen2023text}, Fantasia3D~\cite{chen2023fantasia3d} and Wonder3D~\cite{long2023wonder3d}. Specifically, DreamFusion~\cite{poole2022dreamfusion}, Fantasia3D~\cite{chen2023fantasia3d} and ProlificDreamer~\cite{wang2023prolificdreamer} adopt a similar approach to optimize 3D representation through the score function of a 2D diffusion model, without intervening in 3D priors. We compare our results with these three methods, highlighting the distinct advantages of inferring 3D-consistent geometry and reducing artifacts by incorporating explicit 3D priors. 
Meanwhile, MVDream~\cite{shi2023mvdream} and Wonder3D~\cite{long2023wonder3d} are very recent proposals to use multi-view consistency priors, which derived from finetuned multi-view diffusion models trained on synthetic multi-view rendering image data. GSGEN~\cite{chen2023text}, on the other hand, addresses 3D inconsistency by initializing geometry with Point-E~\cite{nichol2022point} generated shapes. By comparing these three methods, we demonstrate that our introduced 3D priors offer greater generality in challenging and uncommon cases and effectively prevent the generation of 3D assets with lighting and texture styles similar to the synthetic rendered dataset. For DreamFusion~\cite{poole2022dreamfusion}, ProlificDreamer~\cite{wang2023prolificdreamer} and Fantasia3D~\cite{chen2023fantasia3d},  we utilize their implementations in the ThreeStudio~\cite{threestudio2023} library for comparison. For MVDream~\cite{shi2023mvdream}, GSGEN~\cite{chen2023text} and Wonder3D~\cite{long2023wonder3d}, we use their official implementation.

\begin{figure*}[tb]
    \centering
    \includegraphics[width=1\textwidth]{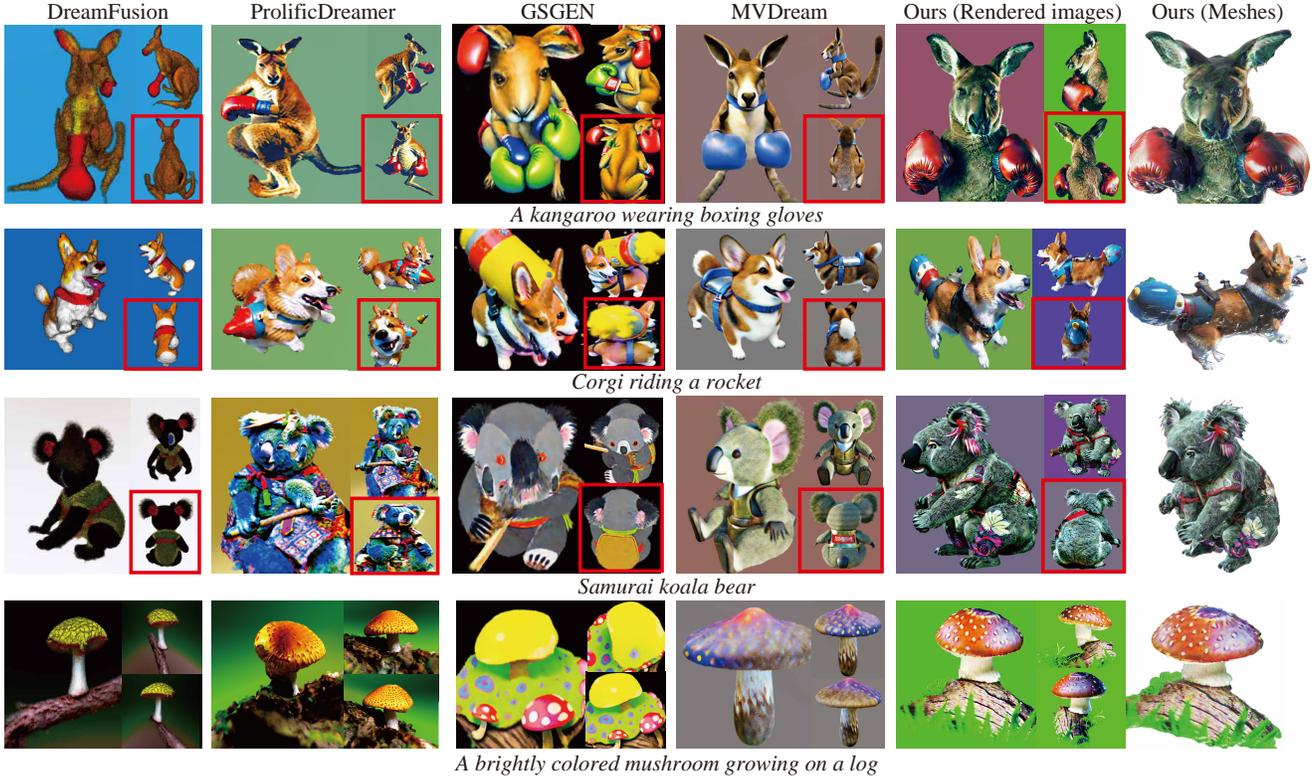}
    \vspace{-0.75cm}
    \caption{Qualitative comparison with baselines. Back views are highlighted with \textcolor{red}{red rectangles} for distinct observation of multiple faces.}
    \label{fig:Qualitative-Comparison}
    \vspace{-0.7cm}
\end{figure*}

\noindent\textbf{Experiment Setup.}
We collected 35 prompts from various sources, including prompts from previous work~\cite{shi2023mvdream,long2023wonder3d} and real user inputs in the wild. To comprehensively assess 3D consistency and semantic coherence, we intentionally selected more prompts indicating asymmetric geometric structures (80\% of the collected prompts) and fewer prompts indicating symmetric geometric structures (20\%). For a fair comparison, we render 3D assets generated by our method and baselines by circling around the object at a default elevation and camera distance~\cite{threestudio2023}, resulting in 120 images. We then evaluate the gap between the rendered images and reference images generated by Stable Diffusion~\cite{rombach2022high} based on the collected prompts. We sample 10$k$ points on the generated meshes to calculate 3D metric. To demonstrate that our method is trivially adaptable to various multi-view diffusion models, we randomly use either Zero123~\cite{liu2023zero} or MVDream~\cite{shi2023mvdream} and Zero123++~\cite{shi2023zero123++} for subsequent experiments. For the effect of different diffusions on the results, please refer to supplementary for detail.

\noindent\textbf{2D Metrics.}
$\rm{FID_{CLIP}}$~\cite{kynkaanniemi2022role} for image fidelity measurement, which is calculated by the disparity in distribution between the rendered image and reference image features, both encoded by CLIP ViT-B-32~\cite{radford2021learning}. CLIP R-score for semantic coherence measurement is calculated by the probability of rendered images retrieving the correct caption among collected prompts. We average the metric over 120 rendered images for the quantitative comparison. 

\noindent\textbf{3D Metric.} These metrics mentioned above are for measuring 2D images. Limited by rendering angles and geometric self-occlusion, 2D metrics often struggle to assess 3D objects in 360 degrees fully. To the best of our knowledge, no metrics have yet been introduced in text-to-3D tasks for evaluating the semantic consistency of 3D assets. Therefore, we propose using Uni3D~\cite{uni3d}, the largest 3D presentation model with one billion parameters under text-image-pointcloud alignment learning objective, to lift semantic coherence measurement from 2D to 3D. We adopt a similar strategy to the CLIP R-score, except that we replace the image and text encoders in the CLIP with the point cloud and text encoders from the Uni3D, referred to as ``$\rm{Uni3D_{score}}$".

\noindent\textbf{Subjective Metric.} 
3D reconstruction tasks are typically evaluated of the error reconstructed shape compared to the ground truth~\cite{NeuralPull}. 
However, these metrics are difficult to apply to text-to-3D tasks, as there is no ground truth. 
We further manually check the number of examples with 3D or semantic inconsistency problems, and then report the rate of success as an auxiliary metric, referred to as ``Cons. Rate".

\noindent\textbf{Quantitative Comparison.}
In Tab.\ref{tab:Qualitative-Comparison1}, we conduct a quantitative comparison over generation quality, text-image consistency and 3D consistency. Overall, the results indicate that our method significantly outperforms the baselines across all metrics, demonstrating that we achieve high-fidelity, text-image and text-3D consistency in the generated quality while ensuring 3D spatial consistency.

\begin{table}
  \caption{Quantitative comparison with baselines.}
  \vspace{-0.3cm}
  \label{tab:Qualitative-Comparison1}
  \centering
  \scalebox{0.67}{
  \begin{tabular}{l|c|cc|c|c}
    \toprule
     \bf{Model}  &  ${\rm{FID}}_{\rm{CLIP}}$ $\downarrow$ &  \multicolumn{2}{c|}{CLIP R-score$\uparrow$}  & ${\rm{Uni3D}}_{\rm{score}}$ $\uparrow$  & Cons. Rate$\uparrow$  \\
    & & \bf {B/16}  & \bf {L/14} & & \\
    \midrule
    DreaFusion~\cite{poole2022dreamfusion}   & 59.6 & 0.844                            &  0.870 & 0.514 & 0.429  \\
    ProlificDreamer~\cite{wang2023prolificdreamer}   & {48.8} & 0.866 & 0.892 & 0.629 & 0.257 \\
    MVDream~\cite{shi2023mvdream} & 50.6 & 0.852 & 0.886 & 0.771 & 0.829  \\
    \hline
    \textbf{Ours} & \textbf{47.9} & \textbf{0.935} & \textbf{0.962} & \textbf{0.800} &\textbf{0.914} \\
    \bottomrule
  \end{tabular}
  }
  \vspace{-0.75cm}
\end{table}

\noindent\textbf{Qualitative Comparison.} Fig.\ref{fig:Qualitative-Comparison} compares our method with the baselines. We present four visual examples: the first three rows depict non-symmetric geometries, while the last row is for symmetric geometry. Notably, we display the front, side, and back views, where the back views are highlighted with red rectangles to enhance the observation of potential multiple faces issues.
We highlight our improvements in visual comparison in Fig.\ref{fig:Qualitative-Comparison}. Dreamfusion and ProlificDreamer produce high-quality frontal views but fail to form a plausible 3D object. In particular, ProlificDreamer delivers photorealistic 3D assets with semantic coherence, where every view resembles canonical views, i.e., the back views that are shown in red rectangles, are mistakenly optimized as front views, resulting in Janus problems. GSGEN mitigates some of the 3D inconsistencies by introducing 3D priors from the pre-trained Point-E. However, the fidelity of the textures it generates is still insufficient for complete satisfaction. Compared to the three methods mentioned above, MVDream stands out as the most effective solution for addressing multi-view inconsistency issues. This is achieved by fine-tuning pre-trained 2D diffusion models using multi-view images rendered from 3D data. Nevertheless, due to the rendering quality and sparsity of 3D training data, the generated results often exhibit cartoon-style textures and semantically lost geometries, particularly when dealing with uncommon and challenging given prompts. For example, it struggles to generate a rocket as required in the second case, a samurai style as required in the third case, and a log as required in the fourth case. 
By incorporating explicit 3D priors with a 2D diffusion model that is capable of imagination diversity, \Ours significantly alleviates the multifaceted nature of generated 3D assets, in terms of both meshes and rendered images exhibiting impressive photorealistic textural details, while maintaining semantic faithfulness, as shown in Fig.\ref{fig:fig1} and Fig.\ref{fig:Qualitative-Comparison}. 
More analysis and comparisons with other baselines can be found in the supplementary.

\begin{figure*}[tb]
    \centering
    \includegraphics[width=1\textwidth]{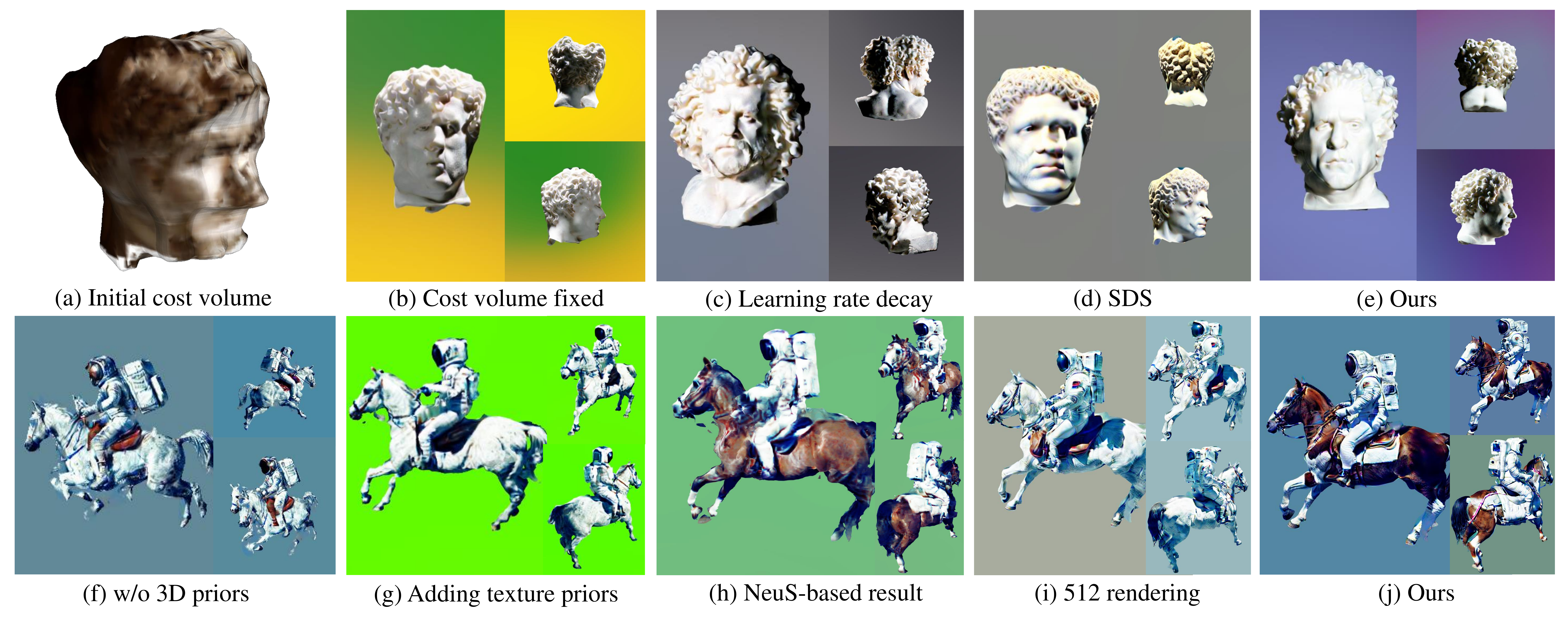}
    \vspace{-0.75cm}
    \caption{Ablation study of proposed improvements for text-to-3D generation.}
    \label{fig:Ablation}
    \vspace{-0.7cm}
\end{figure*}

\subsection{Ablation Study}
\label{sec.ablation}
We then conduct ablation studies to justify the effectiveness of each design in \Ours. We activate all the modules and training strategies mentioned in the Sec.\ref{sec.method} during ablation studies, except for the modified part described in each ablation experiment below. 

\noindent\textbf{The Effect of 3D Priors.} We first visualize the initial cost volume obtained through the volume construction model, as shown in Fig.\ref{fig:Ablation} (a). 
Fig. \ref{fig:Ablation} (a) combined with Fig.\ref{fig:Ablation} (e) demonstrate that relying solely on rough geometric cues can significantly activate potential of 3D awareness in 2D diffusion, alleviating the character's tendency to exhibit multifaceted issues. In contrast to fixed priors in Fig.\ref{fig:Ablation} (b), we propose using optimizable priors that gradually evolve according to the optimization state, thus producing progressively refined results, as shown in Fig.\ref{fig:Ablation} (e) and Fig.\ref{fig:Ablation} (j).
To further assess its impact, we also attempt to deactivate the cost volume, i.e., randomly initializing the 3D prior. The 3D inconsistency issue also arises, as shown in Fig.\ref{fig:Ablation} (f). 
To assess the impact of the learning rate decay schedule, an ablation study is conducted, where the learning rate of the cost volume is set to a suitable constant value. The generated 3D assets still suffer severe degeneration, resulting in a completely collapsed geometry in Fig.\ref{fig:Ablation} (c). The reason is that, during the early stage of optimization, there may be a lot of ambiguity and conflict in the appearance information across different views. Hence, during the early optimization stage, we propose to set the learning rate of the cost volume to a smaller value and gradually increase it for geometric detail optimization. And vice versa for the learning rate of texture, which can prevent content drift in the later stage of optimization, please refer to supplementary for detail.

We further justify whether we should use texture priors. We report a visual result using a pre-trained texture MLP in Sec.\ref{sec.stage1}, rather than reinitializing the MLP network and hash encoding in Sec.\ref{sec.stage2}. Fig.\ref{fig:Ablation} (g) demonstrates that introducing texture priors generally leads to a visual appearance that tends toward non-photorealism and over-smoothing. This observation underlines the necessity of introducing only 3D geometric priors, which only contribute to the geometry modeling during the lifting, avoiding compromising the appearance modeling due to texture priors.

\noindent\textbf{The Effect of Mesh Fine-tuning.} We convert NeuS to DMTet to improve geometric and appearance details. We first show the NeuS-based visual results in Fig.\ref{fig:Ablation} (h). 
\Ours produces better results with finer details, as evidenced in Fig.\ref{fig:Ablation} (j). 
The reason is that the advantages of the 3D assets we generate, which yield improved 3D consistency, lie in the ability to enhance the accuracy of surfaces, thereby reducing the complexity of texture optimization in the DMTet. Fig.\ref{fig:Ablation} (d) presents an ablation study on SDS and VSD loss. SDS is observed to produce over-saturated textures, as opposed to the VSD loss that we default to using. 

\noindent\textbf{The Effect of Rendering Resolution.} Through empirical experimentation, we deduce that collapsed geometry often results in textural distortions, thereby increasing the difficulty of optimization. Hence, we conjecture that 3D consistency is one of the main bottlenecks for increasing the rendering resolution in prior work. Instead, by integrating 3D geometric priors, we achieved better results closer to diffused distributions, making the optimization becomes easier. Consequently, we successfully increase the rendering resolution from 512 to 1024, as shown in Fig.\ref{fig:Ablation} (j). Additionally, Fig.\ref{fig:Ablation} (i) demonstrates that \Ours still provides competitive results at 512 $\times$ 512 resolution.

\section{Conclusion}
We significantly improve the rendering fidelity of images and the details of texture meshes, while greatly alleviating the notorious Janus problems by the awakened 3D-aware capability of 2D diffusion priors, which is unleashed by geometric clues provided by 3D priors in a disentangled solution. Additionally, the disentangled design offers a flexible way to optimize 3D priors gradually. The visual and numerical comparisons with the state-of-the-art methods justify our effectiveness
and show our superiority over the latest methods in 3D generation.

{
    \small
    \bibliographystyle{ieeenat_fullname}
}

\clearpage
\setcounter{page}{1}
\maketitlesupplementary

\section{Video}
Our supplementary material also includes a video, which shows more visualizations, inviting readers to watch for a more intuitive visual experience.

\begin{figure*}[htb]
    \centering
    \includegraphics[width=1\textwidth]{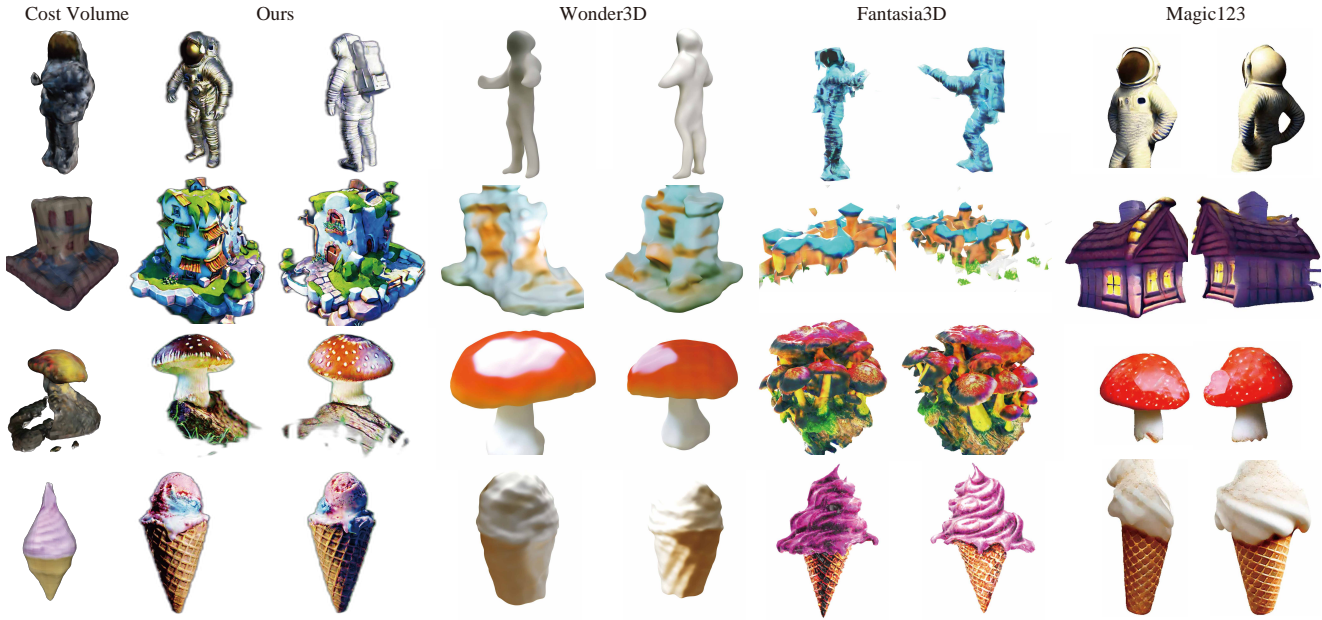}
    \caption{More visualization comparisons with baselines. For each row from up to down, the given prompts are: (1) \emph{3D render of a statue of an astronaut.} (2) \emph{3D stylized game little building.} (3) \emph{A brightly colored mushroom growing on a log.} (4) \emph{An ice-cream cone}}
    \label{fig:more-baselines}
\end{figure*}

\section{Source Code}
To facilitate future research, our code and 3D metric are available at: \href{https://mabaorui.github.io/GeoDream_page}{GeoDream}

\section{Definition of The Janus Problem}
We explain in further detail the definition of the Janus problem (3D inconsistency), which refers to a phenomenon that the learned 3D representation, instead of presenting the 3D desired output, shows multiple canonical views of an object in different directions~\cite{wiki:janus,armandpour2023re}. For instance, when the given prompt indicates an asymmetric geometric structure, such as a person or an animal, the generated 3D asset has multiple faces but lacks complete and correct back views. In contrast, when the given prompt indicates a symmetric structure, such as a cake or a hamburger, which does not have strictly defined back views, issues of 3D inconsistency typically do not arise. Therefore, when calculating the subjective metric, geometrically symmetric 3D assets do not suffer from 3D inconsistency by default.

\section{More Visualization Comparisons with Baselines}
We report our performance with more 3D generation methods, including Fantasia3D~\cite{chen2023fantasia3d}, Wonder3D~\cite{long2023wonder3d}, and Magic123~\cite{lin2023magic3d}. Fantasia3D employs DMTet~\cite{shen2021deep} initialized with a handcrafted 3D model or a predefined geometric shape as the 3D representation, which is the same representation used in our mesh fine-tuning phase. We compare our DMT-based results with Fantasia3D to show the gains in rendering appearance from geometry initialization with 3D priors. Wonder3D employs NeuS~\cite{wang2021neus} as its 3D representation, which is subsequently processed through the Marching Cube algorithm~\cite{lorensen1998marching} to extract mesh. Magic123 adopts a coupled approach, optimizing the 3D representation by using both 3D and 2D priors as losses. The comparisons with Magic123 justify that the disentangling 3D and 2D priors allows for the simultaneous harnessing of the generalization capabilities of 2D diffusion priors and the 3D consistency of 3D priors. In contrast, Magic123 requires careful design of the balance weights between 3D and 2D loss to avoid compromising between the two types of priors. Visual comparisons in Fig.\ref{fig:more-baselines} reveal that we enhance the fidelity and semantic coherence of the generated 3D assets, accompanied by an absence of geometric and textural distortions, indicating excellent 3D spatial consistency.

\begin{figure}[htb]
    \centering
    \includegraphics[width=1\columnwidth]{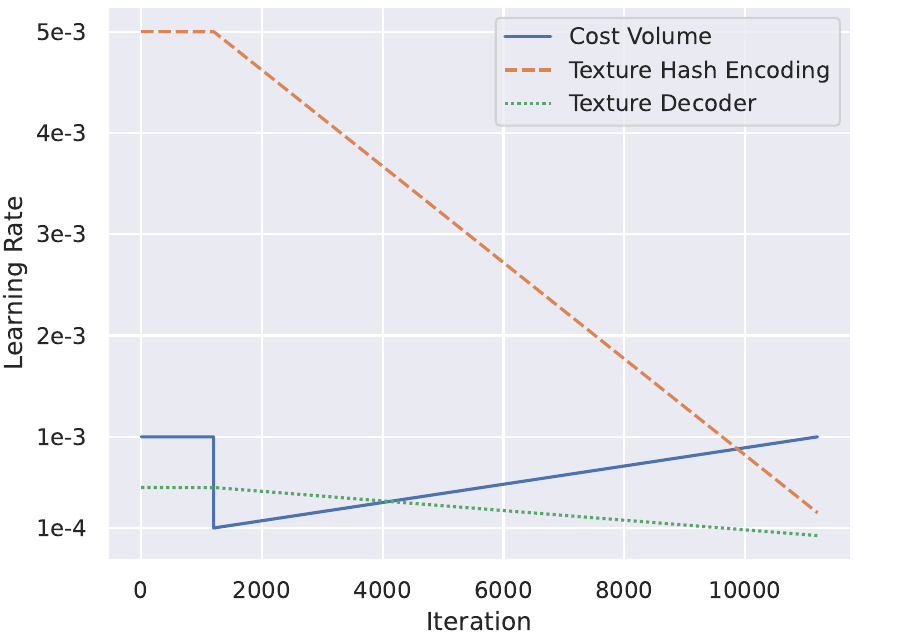}
    \caption{The detailed learning rate schedule.}
    \label{fig:lr}
\end{figure}

\section{Viewpoint Sampling Strategy}
\label{sec.viewsampling}

\begin{figure*}[htb]
    \centering
    \includegraphics[width=1\textwidth]{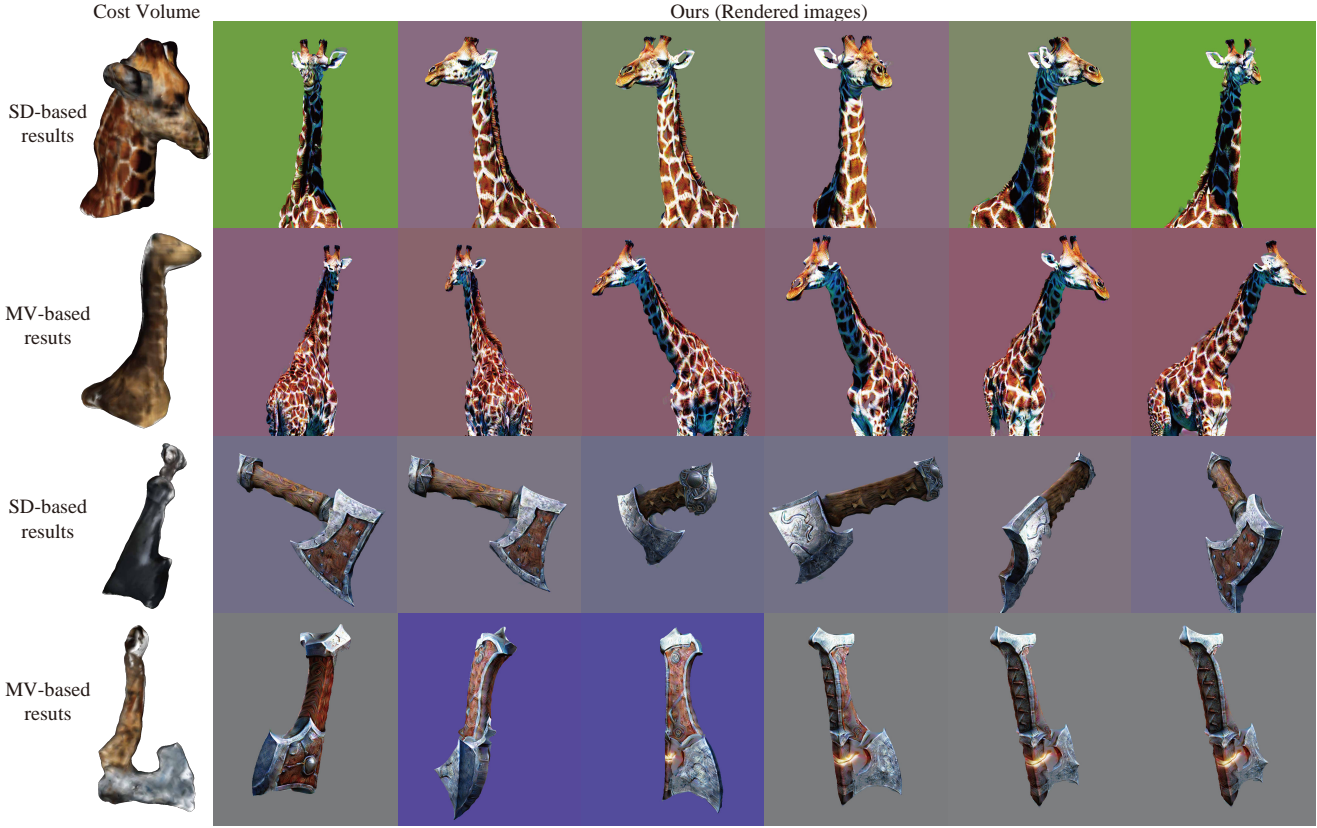}
    \caption{Ablation on the methods for obtaining reference views. We compare the generated 3D assets based on reference views predicted by Stable Diffusion and MVDream, driven by user-provided texts. \Ours adapt to reference views from various sources. For each row from up to down, the given prompts are: (1) \emph{A majestic giraffe with a long neck.} (2) \emph{Viking axe, fantasy, weapon, blender, 8k, HD.}}
    \label{fig:referenceViews}
\end{figure*}

\begin{figure*}[htb]
    \centering
    \includegraphics[width=1\textwidth]{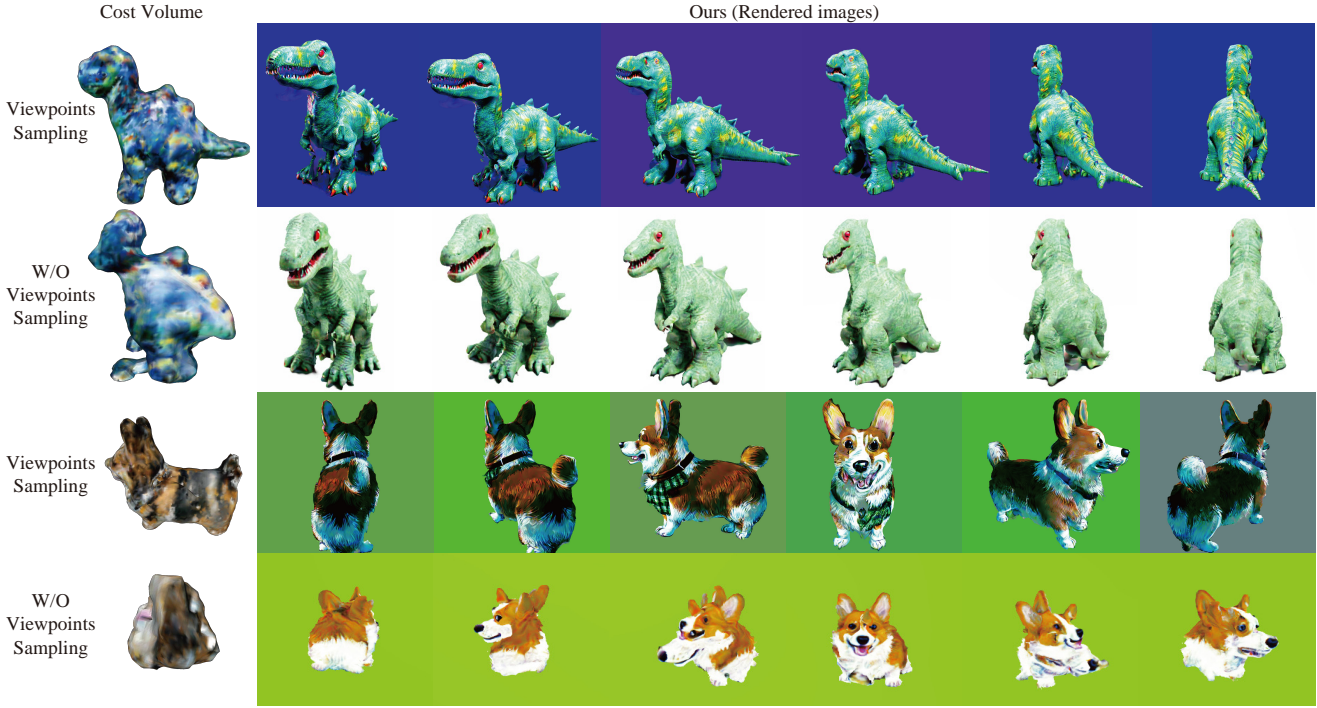}
    \caption{Ablation on the viewpoint sampling strategy. We demonstrate that using our proposed viewpoint sampling strategy contributes to the more robust generation of a consistent cost volume, significantly avoiding the outcomes of geometric collapse. For each row from up to down, the given prompts are: (1) \emph{A dinosaur toy.} (2) \emph{A corgi.}}
    \label{fig:viewSampling}
\end{figure*}

We propose a critical viewpoint sampling strategy to enhance the stability of constructing cost volumes. Cost volume-based methods~\cite{yao2018mvsnet,zhang2022nerfusion,long2022sparseneus,liu2023one} rely on the consistency and accuracy of multi-views to find local correspondences and infer geometry. We empirically find that current multi-view diffusion models~\cite{liu2023zero,yang2023consistnet,liu2023one,li2023sweetdreamer,shi2023mvdream,ye2023consistent} can provide relatively accurate and consistent predictions for small relative pose, when fed with front and side views as reference views. Instead, when a back view is used as the reference view, inconsistencies tend to worsen. Our analysis indicates that these multi-view models are fine-tuned from 2D pre-trained diffusion models, which exhibit weaker performance in predicting non-canonical view information. Additionally, the information implied by back views is quite ambiguous, posing challenges for predicting consistent information. Consequently, we propose a viewpoint sampling strategy to mitigate the aforementioned problems.

Specifically, We obtain reference views driven by a user-provided text in one of two methods: i) Obtaining a front view predicted by Stable Diffusion~\cite{rombach2022high}, which is trivial as Stable Diffusion often biases towards generating canonical views. ii) Utilizing MVDream~\cite{shi2023mvdream} to output desired views based on our predefined absolute camera positions. In our experiments, following the default settings of MVDream, we set the absolute elevation angle at $15^{\circ}$ and absolute azimuth angles at $0^{\circ}$, $90^{\circ}$, $180^{\circ}$, and $270^{\circ}$. We sample four viewpoints on the sphere surface with a default radius to obtain the front, left, back, and right views as reference views.

When the reference view is predicted by Stable Diffusion, we require either Zero123~\cite{liu2023zero} or Zero123++~\cite{shi2023zero123++} to randomly sample viewpoints within a range of a relative azimuth angle less than $180^{\circ}$ and a relative elevation angle less than $30^{\circ}$. Subsequently, we sample an image with a relative azimuth angle of $180^{\circ}$ and a relative elevation angle of $0^{\circ}$ to serve as the back view, which is then added to the source views. In the case of reference views are predicted by MVDream, we use Zero123 or Zero123++ to sample viewpoints relative to the front view, left side view, and right side views, within a range of a relative azimuth angle less than $45^{\circ}$ and a relative elevation angle less than $30^{\circ}$. Subsequently, the back view predicted by MVDream is supplemented to the source views. We show the visualized comparison of the impact of reference views generated by Stable Diffusion and MVDream on the generated 3D assets, as shown in Fig.\ref{fig:referenceViews}. We report visualized results without viewpoint sampling strategy and the results with viewpoint sampling strategy, as shown in Fig.\ref{fig:viewSampling}. The visualized results indicate that our proposed sampling strategy can adapt to reference views predicted by both Stable Diffusion and MVDream, significantly enhancing the quality of the constructed cost volume and the consistency of the generated 3D assets.

Finally, we observe that due to the inherent lack of perfect consistency between source views, the constructed cost volume is quite rough, even with the viewpoint sampling strategy, as shown in Fig.\ref{fig:referenceViews} and Fig.\ref{fig:viewSampling}. However, the ultimately generated 3D assets tend to produce rich details and more complete and consistent geometry. This suggests that disentangling 3D and 2D priors is a potentially exciting direction, as it provides a flexible way to further refine 3D priors while maintaining the ability of 3D priors to unleash 2D diffusion priors.

\section{Learning Rate Decay Schedule}
We propose to set the learning rate of the cost volume to a smaller value and gradually increase it for geometric detail optimization, aiming to maintain geometric priori cues in the early stage of optimization. And vice versa for the learning rate of texture, which can prevent content drift in the later stage of optimization. During the early optimization stage, we adopt an initially high learning rate to fight early overfitting~\cite{li2019towards,he2019control}. The detailed learning rate curves are depicted in Fig.\ref{fig:lr}.

\begin{figure*}[h]
    \centering
    \includegraphics[width=1\textwidth]{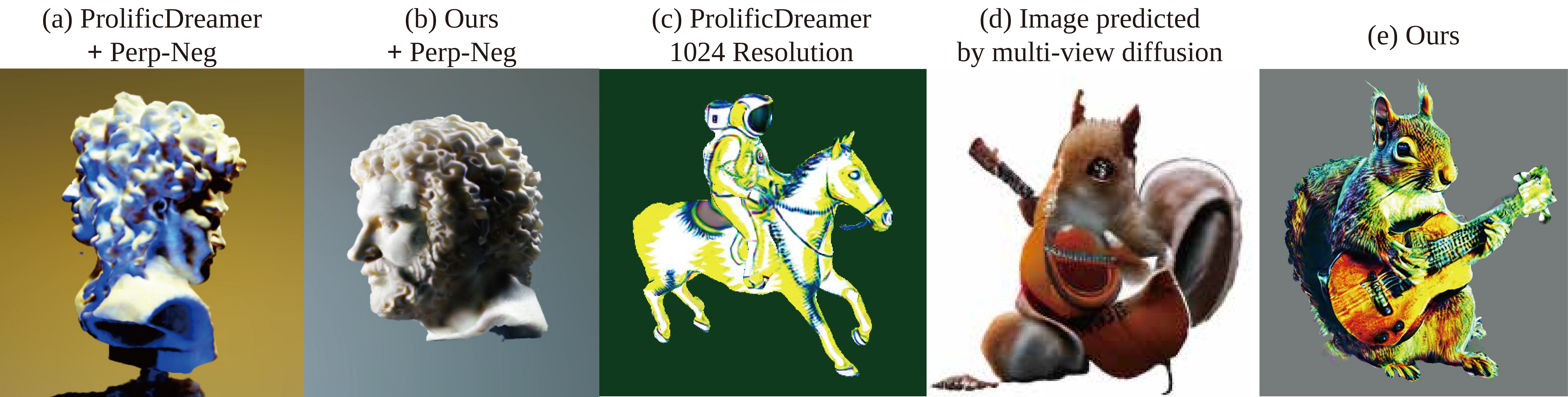}
    \caption{Ablation on negative prompting, rendering resolution, and corner case. The given prompts are: (a) and (b) \emph{A 3D printed white bust of a man with curly hair.} (c) \emph{An astronaut riding a horse.} (d) and (e) \emph{A DSLR photo of a squirrel playing guitar.}}
    \label{fig:prompt}
\end{figure*}

\section{Ablation on negative prompting, rendering resolution, and corner case}

\noindent\textbf{Prompting.} Perp-Neg~\cite{armandpour2023re} introduces a negative prompt algorithm that transforms 2D Diffusion into 3D, addressing the Janus problem. We attempt to integrate the negative prompt algorithm into both ProlificDreamer and \Ours, as shown in Fig.\ref{fig:prompt} (a) and Fig.\ref{fig:prompt} (b). The result shown in Fig.\ref{fig:prompt} (a) demonstrates that the negative prompt algorithm still fails to mitigate the Janus problem stably. Fig.\ref{fig:prompt} (b) illustrates that \Ours is able to yield consistent 3D assets both with and without the negative prompt algorithm. However, since we did not observe a significant improvement in the results, we opt not to use the negative prompt algorithm as a default in our experiments. Instead, we employ view-dependent prompting as in previous works~\cite{poole2022dreamfusion,wang2023prolificdreamer}.

\noindent\textbf{Rendering Resolution.} We attempt to increase the rendering resolution to 1024 in ProlificDreamer, which typically struggles with over-saturation issues, as demonstrated in Fig.\ref{fig:prompt} (c). Our analysis suggests that the absence of 3D priors often leads to collapsed geometry, resulting in textural distortions and thereby increasing the complexity of the optimization.

\noindent\textbf{Corner Case.}
We further explore the robustness of \Ours when faced with failures of multi-view diffusion in predicting multiple views. For instance, when the given prompt is ``\emph{A DSLR photo of a squirrel playing guitar}", multi-view diffusion struggles to accurately predict the correct spatial relationship between the guitar and the squirrel, due to the sparsity of 3D training data. However, \Ours excels in preserving the generalizability and creativity of 2D diffusion priors, enabling more effective compatibility with imperfect multi-view predictions, and thus generating semantically correct 3D assets, as shown in Fig.\ref{fig:prompt} (e).

\section{Training Stability and Diversity}
\begin{figure*}[htb]
    \centering
    \includegraphics[width=1\textwidth]{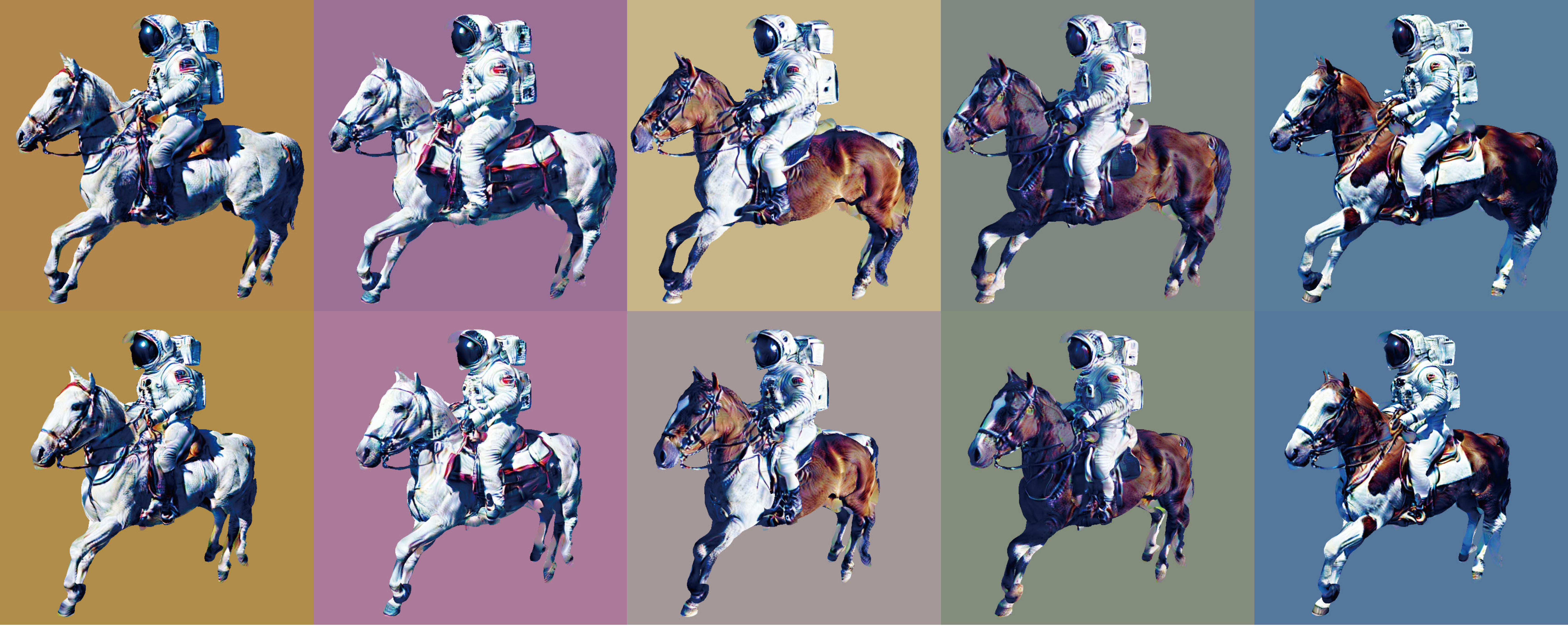}
    \caption{Ablation on training stability. We conduct several experiments on the same prompt to verify the training stability of \Ours. The given prompt is: \emph{An astronaut riding a horse.}}
    \label{fig:Stability}
\end{figure*}

\noindent\textbf{Stability.}
Prior text-to-3D studies are notoriously brittle. The same hyperparameter settings often lead to vastly different results in terms of complete failure or success, depending on the random seed, making them hard to control. To assess the training stability of \Ours, we conduct several experiments on the same prompt, as shown in Fig.\ref{fig:Stability}. \Ours exhibits exceptional training stability. The reason lies in the 3D priors we introduced, which significantly reduce the randomness caused by the random seeds.

\noindent\textbf{Diversity}
Additionally, we can generate diverse 3D models by controlling and leveraging the diversity capabilities of Stable Diffusion or MVDream to predict various reference views, as mentioned in Sec.\ref{sec.viewsampling} and Fig.\ref{fig:referenceViews}. In summary, \Ours provides a balanced solution between diversity and stability.

\section{Licenses}
We provide the URL, citations, and licenses of the open-sourced assets we used in this work, as shown in Tab.\ref{tab:URL}.

\section{Algorithm}
We provide a summarized algorithm of priors refinement in Algorithm \ref{alg:1}.

\section{Training Details}
We construct a cost volume with 150 $\times$ 150 $\times$ 150 voxels in 2 minutes on an NVIDIA-V100-32GB GPU. During the priors refinement stage, we employ a network modified based on ProlificDreamer~\cite{wang2023prolificdreamer}. We replace the learnable hash encoding used in ProlificDreamer by cost volume. We choose a single-layer MLP to decode the color from texture hash encoding as Instant-NGP~\cite{muller2022instant}. Following ProlificDreamer, we set the particle to 1 and utilize v-prediction~\cite{salimans2022progressive} to train the LoRA~\cite{hu2021lora} based on Stable Diffusion v2.1 model for VSD loss. Notably, even when the rendering resolution increased from 512 to 1024, the training time does not show a significant difference compared to ProlificDreamer. The reason is that 3D assets generated by \Ours, which exhibit fewer artifacts and thus enhanced rendering efficiency. Specifically, training the NeuS representation~\cite{wang2021neus} with the batch size set to 1 typically requires approximately 3 hours on a single NVIDIA-V100-32GB GPU. Mesh finetuning with a batch size of 2 usually requires around 8 hours on a single NVIDIA-V100-32GB GPU. Utilizing larger batch sizes and parallel multi-GPUs training could potentially reduce training times and we leave this exploration in future work.

\section{Ablation on Source Views Predicted by Different Multi-View Diffusion Models}
\begin{figure*}[htb]
    \centering
    \includegraphics[width=0.95\textwidth]{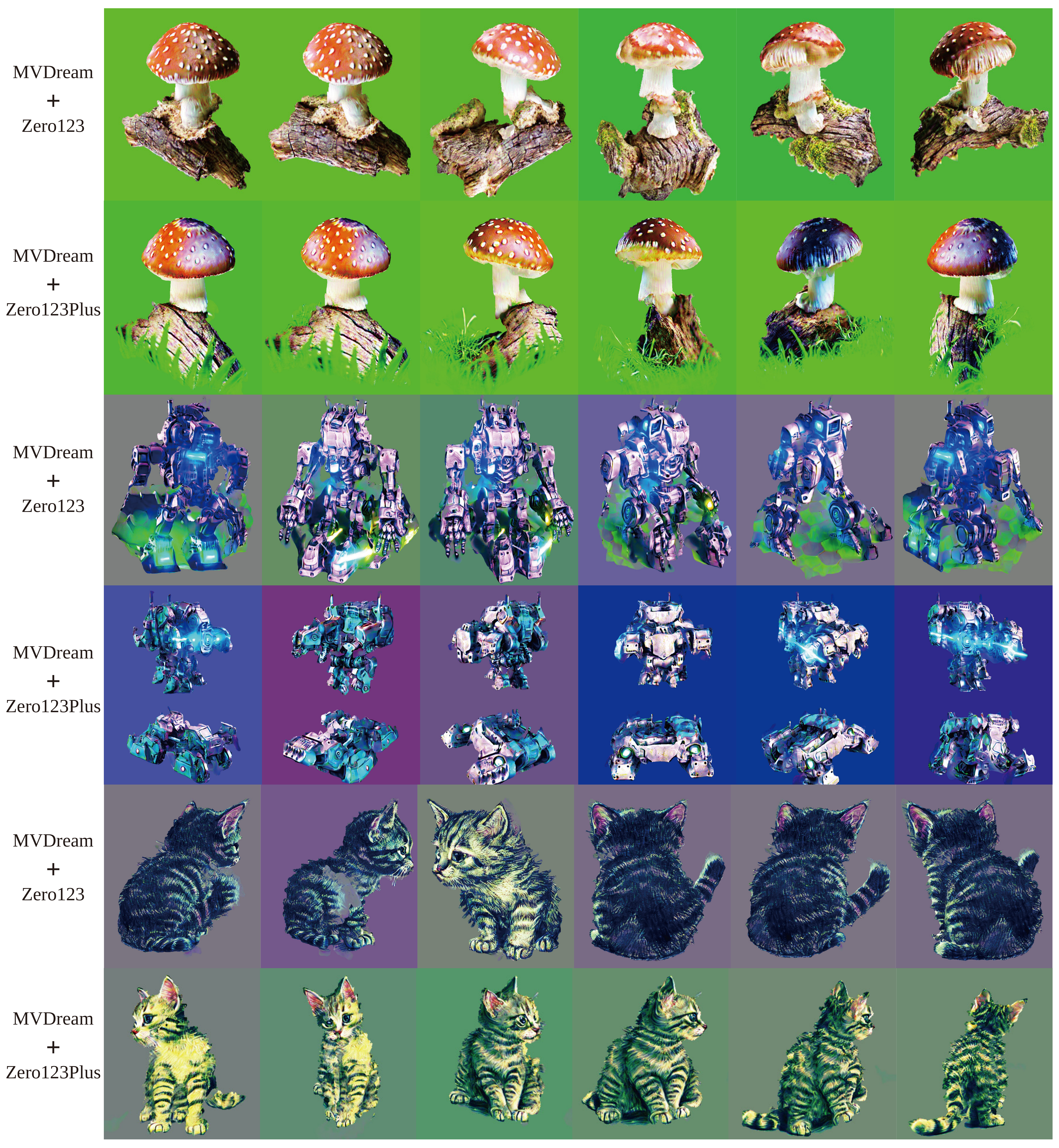}
    \vspace{-0.3cm}
    \caption{Ablation on source views predicted by different multi-view diffusion models. We compare our generated 3D assets based on source views predicted by Zero123 and Zero123++. For a fair comparison, the reference views are generated by MVDream driven by user-provided texts. \Ours adapt to source views predicted by  various multi-view diffusion models. For each row from up to down, the given prompts are: (1) \emph{A brightly colored mushroom growing on a log.} (2) \emph{Mech robot with large weapons on top with hexagonal bases.} (3) \emph{A small kitten.}}
    \label{fig:diff-mv-2}
\end{figure*}

\begin{figure*}[htb]
    \centering
    \includegraphics[width=0.95\textwidth]{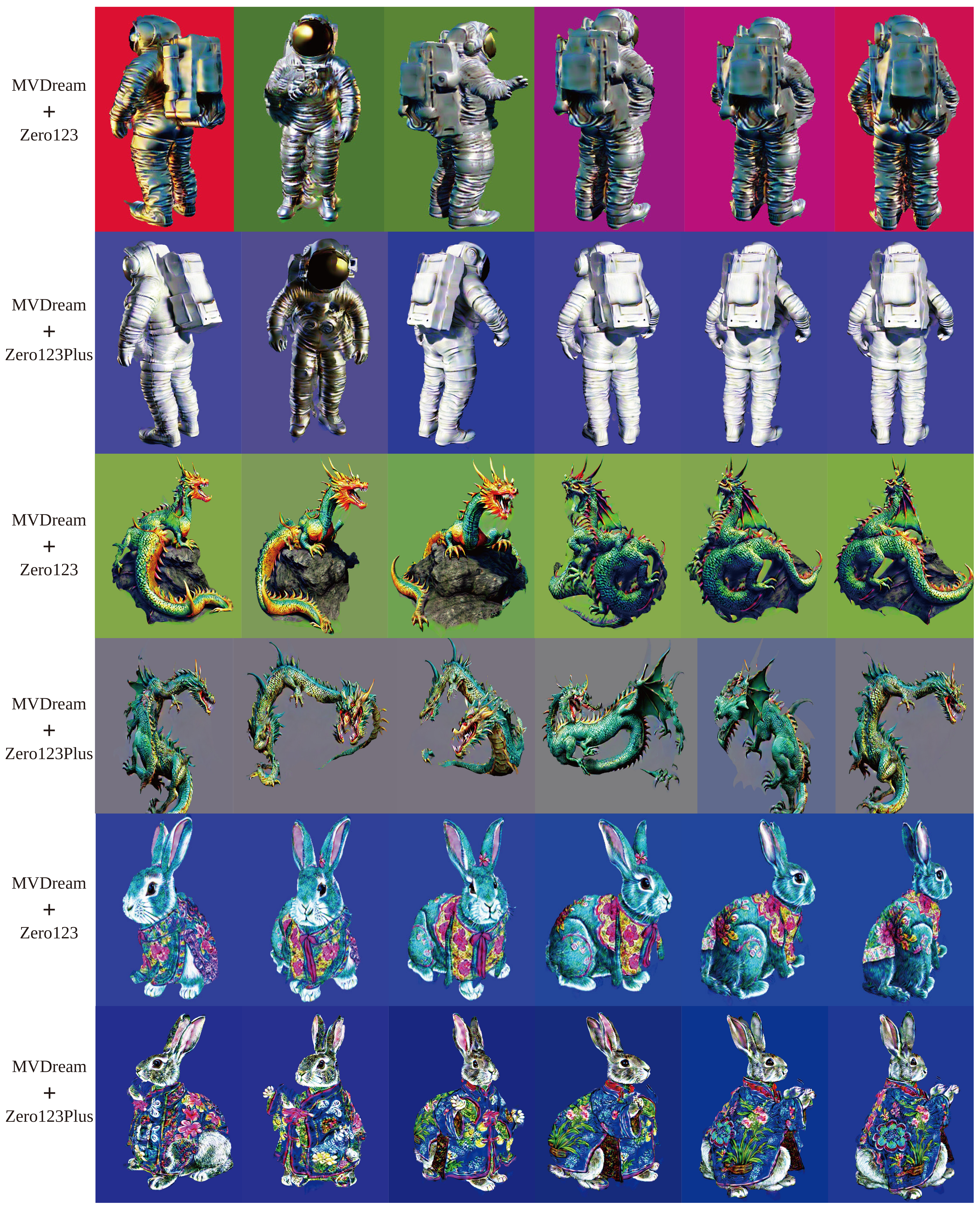}
    \vspace{-0.3cm}
    \caption{Ablation on source views predicted by different multi-view diffusion models. We compare our generated 3D assets based on source views predicted by Zero123 and Zero123++. For a fair comparison, the reference views are generated by MVDream driven by user-provided texts. \Ours adapt to source views predicted by  various multi-view diffusion models. For each row from up to down, the given prompts are: (1) \emph{3D render of a statue of an astronaut.} (2) \emph{A high quality photo of a dragon.} (3) \emph{A cute rabbit in a stunning, detailed Chinese coat.}}
    \label{fig:diff-mv-1}
\end{figure*}

To demonstrate that \Ours is trivially adaptable to various multi-view diffusion models, we conduct the visual comparison with our generated 3D assets based on either Zero123 or Zero123++. Specifically, for a fair comparison, the reference views are generated by MVDream driven by user-provided texts. Then, employing the viewpoint sampling strategy proposed in Sec.\ref{sec.viewsampling}, we obtain source views predicted by Zero123 or Zero123++.
Fig.\ref{fig:diff-mv-2} and Fig.\ref{fig:diff-mv-1} show the comparison of our generated 3D assets based on source views predicted by Zero123 and Zero123++. Fig.\ref{fig:diff-mv-2} and Fig.\ref{fig:diff-mv-1} illustrate that \Ours can adapt to different multi-view diffusion models, producing 3D assets with plausible geometry and intricate rendering details in visual appearance. The adaptability and seamless integration of \Ours with various multi-view diffusion models highlight the evolutionary potential of \Ours, alongside the future advancements of multi-view diffusion models.

\begin{table*}
  \caption{URL, citations and licenses of the open-sourced assets we used in this work.}
  \label{tab:URL}
  \centering
  \setlength{\tabcolsep}{0mm}{
  \begin{tabular}{l c c}
    \hline
    \bf URL & \bf Citation & \bf License \\
    \midrule
    https://github.com/threestudio-project/threestudio & \cite{threestudio2023} & Apache License 2.0 \\
    https://github.com/bytedance/MVDream & \cite{shi2023mvdream} & Apache License 2.0 \\
    https://github.com/One-2-3-45/One-2-3-45 & \cite{liu2023one} & Apache License 2.0 \\
    https://github.com/cvlab-columbia/zero123 & \cite{liu2023zero} & MIT License \\
    https://github.com/SUDO-AI-3D/zero123plus & \cite{shi2023zero123++} & Apache License 2.0 \\
    https://github.com/huggingface/diffusers & \cite{rombach2022high} & Apache License 2.0 \\
    https://github.com/allenai/objaverse-xl & \cite{deitke2023objaverse-xl,deitke2023objaverse} & Apache License 2.0 \\
    \hline
  \end{tabular}
  }
\end{table*}

\begin{algorithm*}
\SetAlgoNlRelativeSize{-1}
\caption{Priors Refinement}
\label{alg:1}
\KwIn{A condition $c$, rotation and translation matrix $\{(R_{i},T_{i})^{N-1}_{i=0}\}$, voxel location $h$, the variance operation $\mathrm{~Var}\{\cdot\}$, the projection procedure $P(\cdot,\cdot)$, multi-view diffusion $f_{mv}$, a 2D feature network $f_{2D}$, a 3D feature network $f_{3D}$, a geometric decoder $f_{g}$, texture decoder $f_{t}'$ , position encoding $E(\cdot)$, 2D diffusion model $\epsilon_{pretrain}$. Learning rate $\eta_{1}$, $\eta_{2}$, $\eta_{3}$,$\eta_{4}$ and $\eta_{5}$ for cost volume $V$, hash texture encoding $h_{\Omega}$, texture decoder $f_{t}'$, a LoRA diffusion model $\epsilon_{l}$ and DMTet parameters, respectively.}

Initialize 2D feature network $f_{2D}$, 3D feature network $f_{3D}$, and geometry MLP decoder $f_{g}$ with pretrained parameters obtained from 3D priors training stage. Initialize texture hash encoding and texture decoder $f_{t}'$ parameterized by $({\theta}_{2}, {\theta}_{3})$. Initialize a LoRA diffusion model parameterized by  $l$.

\For{i=0 to N-1}{
$F^p_{i} \leftarrow f_{2D}(f_{mv}(c,R_{i},T_{i}))$
}

$V_p = f_{3D}(\mathrm{~Var}\{P(F^p_{i},  h)\}_{i=0}^{N-1})$

Cost volume $V_p$ parameterized by ${\theta}_{1}$.

\While{not converged}{
    Ramdomly sample a camera pose $o$. Sample $M$ query points $x_j$ along the view ray based on camera pose $o$.

    \For{j=0 to M-1}{
    $s_j \leftarrow \,f_{g}(E(x_j), V_P(x_j))$

    $c_j \leftarrow f_{t}'(h_{\Omega}(x_j), x_j)$
    }
    
    $\hat{x} \leftarrow R(\{s_j\}_{j=0}^{M-1}, \{c_j\}_{j=0}^{M-1})$
    
    ${\theta}_{1} \leftarrow {\theta}_{1} - \eta_{1} \mathrm{E}_{t, \epsilon, o} [w(t)(\epsilon_{pretrain}(\hat{x_{t}}, t, c) - \epsilon_{l}(\hat{x_{t}}, t, c, o))\frac{\partial \hat{x}}{\partial {\theta}_{1}}]$

    ${\theta}_{2} \leftarrow {\theta}_{2} - \eta_{2} \mathrm{E}_{t, \epsilon, o} [w(t)(\epsilon_{pretrain}(\hat{x_{t}}, t, c) - \epsilon_{l}(\hat{x_{t}}, t, c, o))\frac{\partial \hat{x}}{\partial {\theta}_{2}}]$

    ${\theta}_{3} \leftarrow {\theta}_{3} - \eta_{3} \mathrm{E}_{t, \epsilon, o} [w(t)(\epsilon_{pretrain}(\hat{x_{t}}, t, c) - \epsilon_{l}(\hat{x_{t}}, t, c, o))\frac{\partial \hat{x}}{\partial {\theta}_{3}}]$

    $l \leftarrow l - \eta_{4} \nabla_{l} \mathrm{E}_{t, \epsilon} ||\epsilon_{l}(\hat{x_{t}}, t, c, o)) - \epsilon||^{2}_{2}$
}

Mesh fine-tuning, we use DMTet to extract textured mesh from optimized 3D representation parameterized by $({\theta}_{1}, {\theta}_{2},{\theta}_{3})$ and geometry MLP decoder $f_{g}$. The extracted DMTet parameterized by ${\theta}_{4}$. Initialize a LoRA diffusion model parameters $l'$.

\While{not converged}{
    Ramdomly sample a camera pose $o$. Render 2D image $\hat{x}$ at pose $o$.

    ${\theta}_{5} \leftarrow {\theta}_{5} - \eta_{5} \mathrm{E}_{t, \epsilon, o} [w(t)(\epsilon_{pretrain}(\hat{x_{t}}, t, c) - \epsilon_{l'}(\hat{x_{t}}, t, c, o))\frac{\partial \hat{x}}{\partial {\theta}_{5}}]$

    $l' \leftarrow l' - \eta_{4} \nabla_{l'} \mathrm{E}_{t, \epsilon} ||\epsilon_{l'}(\hat{x_{t}}, t, c, o)) - \epsilon||^{2}_{2}$
}

\Return{}
\end{algorithm*}

\cleardoublepage
\bibliography{main}

\begin{thebibliography}{60}
\providecommand{\natexlab}[1]{#1}
\providecommand{\url}[1]{\texttt{#1}}
\expandafter\ifx\csname urlstyle\endcsname\relax
  \providecommand{\doi}[1]{doi: #1}\else
  \providecommand{\doi}{doi: \begingroup \urlstyle{rm}\Url}\fi

\bibitem[Armandpour et~al.(2023)Armandpour, Zheng, Sadeghian, Sadeghian, and Zhou]{armandpour2023re}
Mohammadreza Armandpour, Huangjie Zheng, Ali Sadeghian, Amir Sadeghian, and Mingyuan Zhou.
\newblock Re-imagine the negative prompt algorithm: Transform 2d diffusion into 3d, alleviate janus problem and beyond.
\newblock \emph{arXiv preprint arXiv:2304.04968}, 2023.

\bibitem[Chan et~al.(2022)Chan, Lin, Chan, Nagano, Pan, De~Mello, Gallo, Guibas, Tremblay, Khamis, et~al.]{chan2022efficient}
Eric~R Chan, Connor~Z Lin, Matthew~A Chan, Koki Nagano, Boxiao Pan, Shalini De~Mello, Orazio Gallo, Leonidas~J Guibas, Jonathan Tremblay, Sameh Khamis, et~al.
\newblock Efficient geometry-aware 3d generative adversarial networks.
\newblock In \emph{Proceedings of the IEEE/CVF Conference on Computer Vision and Pattern Recognition}, pages 16123--16133, 2022.

\bibitem[Chen et~al.(2023{\natexlab{a}})Chen, Chen, Jiao, and Jia]{chen2023fantasia3d}
Rui Chen, Yongwei Chen, Ningxin Jiao, and Kui Jia.
\newblock Fantasia3d: Disentangling geometry and appearance for high-quality text-to-3d content creation.
\newblock \emph{arXiv preprint arXiv:2303.13873}, 2023{\natexlab{a}}.

\bibitem[Chen et~al.(2023{\natexlab{b}})Chen, Wang, and Liu]{chen2023text}
Zilong Chen, Feng Wang, and Huaping Liu.
\newblock Text-to-3d using gaussian splatting.
\newblock \emph{arXiv preprint arXiv:2309.16585}, 2023{\natexlab{b}}.

\bibitem[Deitke et~al.(2023{\natexlab{a}})Deitke, Liu, Wallingford, Ngo, Michel, Kusupati, Fan, Laforte, Voleti, Gadre, et~al.]{deitke2023objaverse-xl}
Matt Deitke, Ruoshi Liu, Matthew Wallingford, Huong Ngo, Oscar Michel, Aditya Kusupati, Alan Fan, Christian Laforte, Vikram Voleti, Samir~Yitzhak Gadre, et~al.
\newblock Objaverse-xl: A universe of 10m+ 3d objects.
\newblock \emph{arXiv preprint arXiv:2307.05663}, 2023{\natexlab{a}}.

\bibitem[Deitke et~al.(2023{\natexlab{b}})Deitke, Schwenk, Salvador, Weihs, Michel, VanderBilt, Schmidt, Ehsani, Kembhavi, and Farhadi]{deitke2023objaverse}
Matt Deitke, Dustin Schwenk, Jordi Salvador, Luca Weihs, Oscar Michel, Eli VanderBilt, Ludwig Schmidt, Kiana Ehsani, Aniruddha Kembhavi, and Ali Farhadi.
\newblock Objaverse: A universe of annotated 3d objects.
\newblock In \emph{Proceedings of the IEEE/CVF Conference on Computer Vision and Pattern Recognition}, pages 13142--13153, 2023{\natexlab{b}}.

\bibitem[Deng et~al.(2022)Deng, Yang, Xiang, and Tong]{deng2022gram}
Yu Deng, Jiaolong Yang, Jianfeng Xiang, and Xin Tong.
\newblock Gram: Generative radiance manifolds for 3d-aware image generation.
\newblock In \emph{Proceedings of the IEEE/CVF Conference on Computer Vision and Pattern Recognition}, pages 10673--10683, 2022.

\bibitem[Goodfellow et~al.(2014)Goodfellow, Pouget-Abadie, Mirza, Xu, Warde-Farley, Ozair, Courville, and Bengio]{goodfellow2014generative}
Ian Goodfellow, Jean Pouget-Abadie, Mehdi Mirza, Bing Xu, David Warde-Farley, Sherjil Ozair, Aaron Courville, and Yoshua Bengio.
\newblock Generative adversarial nets.
\newblock \emph{Advances in neural information processing systems}, 27, 2014.

\bibitem[Guo et~al.(2023)Guo, Liu, Shao, Laforte, Voleti, Luo, Chen, Zou, Wang, Cao, and Zhang]{threestudio2023}
Yuan-Chen Guo, Ying-Tian Liu, Ruizhi Shao, Christian Laforte, Vikram Voleti, Guan Luo, Chia-Hao Chen, Zi-Xin Zou, Chen Wang, Yan-Pei Cao, and Song-Hai Zhang.
\newblock threestudio: A unified framework for 3d content generation.
\newblock \url{https://github.com/threestudio-project/threestudio}, 2023.

\bibitem[He et~al.(2019)He, Liu, and Tao]{he2019control}
Fengxiang He, Tongliang Liu, and Dacheng Tao.
\newblock Control batch size and learning rate to generalize well: Theoretical and empirical evidence.
\newblock \emph{Advances in neural information processing systems}, 32, 2019.

\bibitem[Henderson and Ferrari(2020)]{henderson2020learning}
Paul Henderson and Vittorio Ferrari.
\newblock Learning single-image 3d reconstruction by generative modelling of shape, pose and shading.
\newblock \emph{International Journal of Computer Vision}, 128\penalty0 (4):\penalty0 835--854, 2020.

\bibitem[Henderson et~al.(2020)Henderson, Tsiminaki, and Lampert]{henderson2020leveraging}
Paul Henderson, Vagia Tsiminaki, and Christoph~H Lampert.
\newblock Leveraging 2d data to learn textured 3d mesh generation.
\newblock In \emph{Proceedings of the IEEE/CVF conference on computer vision and pattern recognition}, pages 7498--7507, 2020.

\bibitem[Hong et~al.(2023)Hong, Ahn, and Kim]{hong2023debiasing}
Susung Hong, Donghoon Ahn, and Seungryong Kim.
\newblock Debiasing scores and prompts of 2d diffusion for robust text-to-3d generation.
\newblock \emph{arXiv preprint arXiv:2303.15413}, 2023.

\bibitem[Hu et~al.(2021)Hu, Shen, Wallis, Allen-Zhu, Li, Wang, Wang, and Chen]{hu2021lora}
Edward~J Hu, Yelong Shen, Phillip Wallis, Zeyuan Allen-Zhu, Yuanzhi Li, Shean Wang, Lu Wang, and Weizhu Chen.
\newblock Lora: Low-rank adaptation of large language models.
\newblock \emph{arXiv preprint arXiv:2106.09685}, 2021.

\bibitem[Jun and Nichol(2023)]{jun2023shap}
Heewoo Jun and Alex Nichol.
\newblock Shap-e: Generating conditional 3d implicit functions.
\newblock \emph{arXiv preprint arXiv:2305.02463}, 2023.

\bibitem[Karnewar et~al.(2023)Karnewar, Vedaldi, Novotny, and Mitra]{karnewar2023holodiffusion}
Animesh Karnewar, Andrea Vedaldi, David Novotny, and Niloy~J Mitra.
\newblock Holodiffusion: Training a 3d diffusion model using 2d images.
\newblock In \emph{Proceedings of the IEEE/CVF Conference on Computer Vision and Pattern Recognition}, pages 18423--18433, 2023.

\bibitem[Kingma and Welling(2013)]{kingma2013auto}
Diederik~P Kingma and Max Welling.
\newblock Auto-encoding variational bayes.
\newblock \emph{arXiv preprint arXiv:1312.6114}, 2013.

\bibitem[Kynk{\"a}{\"a}nniemi et~al.(2022)Kynk{\"a}{\"a}nniemi, Karras, Aittala, Aila, and Lehtinen]{kynkaanniemi2022role}
Tuomas Kynk{\"a}{\"a}nniemi, Tero Karras, Miika Aittala, Timo Aila, and Jaakko Lehtinen.
\newblock The role of imagenet classes in fr$\backslash$'echet inception distance.
\newblock \emph{arXiv preprint arXiv:2203.06026}, 2022.

\bibitem[Li et~al.(2023{\natexlab{a}})Li, Duan, Zhou, and Lu]{li2023diffusion}
Muheng Li, Yueqi Duan, Jie Zhou, and Jiwen Lu.
\newblock Diffusion-sdf: Text-to-shape via voxelized diffusion.
\newblock In \emph{Proceedings of the IEEE/CVF Conference on Computer Vision and Pattern Recognition}, pages 12642--12651, 2023{\natexlab{a}}.

\bibitem[Li et~al.(2023{\natexlab{b}})Li, Chen, Chen, and Tan]{li2023sweetdreamer}
Weiyu Li, Rui Chen, Xuelin Chen, and Ping Tan.
\newblock Sweetdreamer: Aligning geometric priors in 2d diffusion for consistent text-to-3d.
\newblock \emph{arXiv preprint arXiv:2310.02596}, 2023{\natexlab{b}}.

\bibitem[Li et~al.(2019)Li, Wei, and Ma]{li2019towards}
Yuanzhi Li, Colin Wei, and Tengyu Ma.
\newblock Towards explaining the regularization effect of initial large learning rate in training neural networks.
\newblock \emph{Advances in Neural Information Processing Systems}, 32, 2019.

\bibitem[Lin et~al.(2023)Lin, Gao, Tang, Takikawa, Zeng, Huang, Kreis, Fidler, Liu, and Lin]{lin2023magic3d}
Chen-Hsuan Lin, Jun Gao, Luming Tang, Towaki Takikawa, Xiaohui Zeng, Xun Huang, Karsten Kreis, Sanja Fidler, Ming-Yu Liu, and Tsung-Yi Lin.
\newblock Magic3d: High-resolution text-to-3d content creation.
\newblock In \emph{Proceedings of the IEEE/CVF Conference on Computer Vision and Pattern Recognition}, pages 300--309, 2023.

\bibitem[Liu et~al.(2023{\natexlab{a}})Liu, Xu, Jin, Chen, Xu, Su, et~al.]{liu2023one}
Minghua Liu, Chao Xu, Haian Jin, Linghao Chen, Zexiang Xu, Hao Su, et~al.
\newblock One-2-3-45: Any single image to 3d mesh in 45 seconds without per-shape optimization.
\newblock \emph{arXiv preprint arXiv:2306.16928}, 2023{\natexlab{a}}.

\bibitem[Liu et~al.(2023{\natexlab{b}})Liu, Wu, Van~Hoorick, Tokmakov, Zakharov, and Vondrick]{liu2023zero}
Ruoshi Liu, Rundi Wu, Basile Van~Hoorick, Pavel Tokmakov, Sergey Zakharov, and Carl Vondrick.
\newblock Zero-1-to-3: Zero-shot one image to 3d object.
\newblock In \emph{Proceedings of the IEEE/CVF International Conference on Computer Vision}, pages 9298--9309, 2023{\natexlab{b}}.

\bibitem[Liu et~al.(2023{\natexlab{c}})Liu, Lin, Zeng, Long, Liu, Komura, and Wang]{liu2023syncdreamer}
Yuan Liu, Cheng Lin, Zijiao Zeng, Xiaoxiao Long, Lingjie Liu, Taku Komura, and Wenping Wang.
\newblock Syncdreamer: Generating multiview-consistent images from a single-view image.
\newblock \emph{arXiv preprint arXiv:2309.03453}, 2023{\natexlab{c}}.

\bibitem[Long et~al.(2022)Long, Lin, Wang, Komura, and Wang]{long2022sparseneus}
Xiaoxiao Long, Cheng Lin, Peng Wang, Taku Komura, and Wenping Wang.
\newblock Sparseneus: Fast generalizable neural surface reconstruction from sparse views.
\newblock In \emph{European Conference on Computer Vision}, pages 210--227. Springer, 2022.

\bibitem[Long et~al.(2023)Long, Guo, Lin, Liu, Dou, Liu, Ma, Zhang, Habermann, Theobalt, et~al.]{long2023wonder3d}
Xiaoxiao Long, Yuan-Chen Guo, Cheng Lin, Yuan Liu, Zhiyang Dou, Lingjie Liu, Yuexin Ma, Song-Hai Zhang, Marc Habermann, Christian Theobalt, et~al.
\newblock Wonder3d: Single image to 3d using cross-domain diffusion.
\newblock \emph{arXiv preprint arXiv:2310.15008}, 2023.

\bibitem[Lorensen and Cline(1998)]{lorensen1998marching}
William~E Lorensen and Harvey~E Cline.
\newblock Marching cubes: A high resolution 3d surface construction algorithm.
\newblock In \emph{Seminal graphics: pioneering efforts that shaped the field}, pages 347--353. 1998.

\bibitem[Ma et~al.(2021)Ma, Han, Liu, and Zwicker]{NeuralPull}
Baorui Ma, Zhizhong Han, Yu-Shen Liu, and Matthias Zwicker.
\newblock Neural-pull: Learning signed distance functions from point clouds by learning to pull space onto surfaces.
\newblock In \emph{International Conference on Machine Learning (ICML)}, 2021.

\bibitem[Melas-Kyriazi et~al.(2023)Melas-Kyriazi, Laina, Rupprecht, and Vedaldi]{melas2023realfusion}
Luke Melas-Kyriazi, Iro Laina, Christian Rupprecht, and Andrea Vedaldi.
\newblock Realfusion: 360deg reconstruction of any object from a single image.
\newblock In \emph{Proceedings of the IEEE/CVF Conference on Computer Vision and Pattern Recognition}, pages 8446--8455, 2023.

\bibitem[Mildenhall et~al.(2021)Mildenhall, Srinivasan, Tancik, Barron, Ramamoorthi, and Ng]{mildenhall2021nerf}
Ben Mildenhall, Pratul~P Srinivasan, Matthew Tancik, Jonathan~T Barron, Ravi Ramamoorthi, and Ren Ng.
\newblock Nerf: Representing scenes as neural radiance fields for view synthesis.
\newblock \emph{Communications of the ACM}, 65\penalty0 (1):\penalty0 99--106, 2021.

\bibitem[Mo et~al.(2023)Mo, Xie, Chu, Yao, Hong, Nie{\ss}ner, and Li]{mo2023dit}
Shentong Mo, Enze Xie, Ruihang Chu, Lewei Yao, Lanqing Hong, Matthias Nie{\ss}ner, and Zhenguo Li.
\newblock Dit-3d: Exploring plain diffusion transformers for 3d shape generation.
\newblock \emph{arXiv preprint arXiv:2307.01831}, 2023.

\bibitem[M{\"u}ller et~al.(2022)M{\"u}ller, Evans, Schied, and Keller]{muller2022instant}
Thomas M{\"u}ller, Alex Evans, Christoph Schied, and Alexander Keller.
\newblock Instant neural graphics primitives with a multiresolution hash encoding.
\newblock \emph{ACM Transactions on Graphics (ToG)}, 41\penalty0 (4):\penalty0 1--15, 2022.

\bibitem[Nichol et~al.(2022)Nichol, Jun, Dhariwal, Mishkin, and Chen]{nichol2022point}
Alex Nichol, Heewoo Jun, Prafulla Dhariwal, Pamela Mishkin, and Mark Chen.
\newblock Point-e: A system for generating 3d point clouds from complex prompts.
\newblock \emph{arXiv preprint arXiv:2212.08751}, 2022.

\bibitem[Poole et~al.(2022)Poole, Jain, Barron, and Mildenhall]{poole2022dreamfusion}
Ben Poole, Ajay Jain, Jonathan~T Barron, and Ben Mildenhall.
\newblock Dreamfusion: Text-to-3d using 2d diffusion.
\newblock \emph{arXiv preprint arXiv:2209.14988}, 2022.

\bibitem[Purushwalkam and Naik(2023)]{purushwalkam2023conrad}
Senthil Purushwalkam and Nikhil Naik.
\newblock Conrad: Image constrained radiance fields for 3d generation from a single image.
\newblock In \emph{Thirty-seventh Conference on Neural Information Processing Systems}, 2023.

\bibitem[Qian et~al.(2023)Qian, Mai, Hamdi, Ren, Siarohin, Li, Lee, Skorokhodov, Wonka, Tulyakov, and Ghanem]{qian2023magic123}
Guocheng Qian, Jinjie Mai, Abdullah Hamdi, Jian Ren, Aliaksandr Siarohin, Bing Li, Hsin-Ying Lee, Ivan Skorokhodov, Peter Wonka, Sergey Tulyakov, and Bernard Ghanem.
\newblock Magic123: One image to high-quality 3d object generation using both 2d and 3d diffusion priors.
\newblock \emph{arXiv preprint arXiv:2306.17843}, 2023.

\bibitem[Radford et~al.(2021)Radford, Kim, Hallacy, Ramesh, Goh, Agarwal, Sastry, Askell, Mishkin, Clark, et~al.]{radford2021learning}
Alec Radford, Jong~Wook Kim, Chris Hallacy, Aditya Ramesh, Gabriel Goh, Sandhini Agarwal, Girish Sastry, Amanda Askell, Pamela Mishkin, Jack Clark, et~al.
\newblock Learning transferable visual models from natural language supervision.
\newblock In \emph{International conference on machine learning}, pages 8748--8763. PMLR, 2021.

\bibitem[Ramesh et~al.(2022)Ramesh, Dhariwal, Nichol, Chu, and Chen]{ramesh2022hierarchical}
Aditya Ramesh, Prafulla Dhariwal, Alex Nichol, Casey Chu, and Mark Chen.
\newblock Hierarchical text-conditional image generation with clip latents, 2022.
\newblock \emph{URL https://arxiv. org/abs/2204.06125}, 7, 2022.

\bibitem[Rombach et~al.(2022)Rombach, Blattmann, Lorenz, Esser, and Ommer]{rombach2022high}
Robin Rombach, Andreas Blattmann, Dominik Lorenz, Patrick Esser, and Bj{\"o}rn Ommer.
\newblock High-resolution image synthesis with latent diffusion models.
\newblock In \emph{Proceedings of the IEEE/CVF conference on computer vision and pattern recognition}, pages 10684--10695, 2022.

\bibitem[Saharia et~al.(2022)Saharia, Chan, Saxena, Li, Whang, Denton, Ghasemipour, Gontijo~Lopes, Karagol~Ayan, Salimans, et~al.]{saharia2022photorealistic}
Chitwan Saharia, William Chan, Saurabh Saxena, Lala Li, Jay Whang, Emily~L Denton, Kamyar Ghasemipour, Raphael Gontijo~Lopes, Burcu Karagol~Ayan, Tim Salimans, et~al.
\newblock Photorealistic text-to-image diffusion models with deep language understanding.
\newblock \emph{Advances in Neural Information Processing Systems}, 35:\penalty0 36479--36494, 2022.

\bibitem[Salimans and Ho(2022)]{salimans2022progressive}
Tim Salimans and Jonathan Ho.
\newblock Progressive distillation for fast sampling of diffusion models.
\newblock \emph{arXiv preprint arXiv:2202.00512}, 2022.

\bibitem[Shen et~al.(2021)Shen, Gao, Yin, Liu, and Fidler]{shen2021deep}
Tianchang Shen, Jun Gao, Kangxue Yin, Ming-Yu Liu, and Sanja Fidler.
\newblock Deep marching tetrahedra: a hybrid representation for high-resolution 3d shape synthesis.
\newblock \emph{Advances in Neural Information Processing Systems}, 34:\penalty0 6087--6101, 2021.

\bibitem[Shi et~al.(2023{\natexlab{a}})Shi, Chen, Zhang, Liu, Xu, Wei, Chen, Zeng, and Su]{shi2023zero123++}
Ruoxi Shi, Hansheng Chen, Zhuoyang Zhang, Minghua Liu, Chao Xu, Xinyue Wei, Linghao Chen, Chong Zeng, and Hao Su.
\newblock Zero123++: a single image to consistent multi-view diffusion base model.
\newblock \emph{arXiv preprint arXiv:2310.15110}, 2023{\natexlab{a}}.

\bibitem[Shi et~al.(2023{\natexlab{b}})Shi, Wang, Ye, Long, Li, and Yang]{shi2023mvdream}
Yichun Shi, Peng Wang, Jianglong Ye, Mai Long, Kejie Li, and Xiao Yang.
\newblock Mvdream: Multi-view diffusion for 3d generation.
\newblock \emph{arXiv preprint arXiv:2308.16512}, 2023{\natexlab{b}}.

\bibitem[Shue et~al.(2023)Shue, Chan, Po, Ankner, Wu, and Wetzstein]{shue20233d}
J~Ryan Shue, Eric~Ryan Chan, Ryan Po, Zachary Ankner, Jiajun Wu, and Gordon Wetzstein.
\newblock 3d neural field generation using triplane diffusion.
\newblock In \emph{Proceedings of the IEEE/CVF Conference on Computer Vision and Pattern Recognition}, pages 20875--20886, 2023.

\bibitem[Sun et~al.(2023)Sun, Zhang, Shao, Wang, Liu, Xie, and Liu]{sun2023dreamcraft3d}
Jingxiang Sun, Bo Zhang, Ruizhi Shao, Lizhen Wang, Wen Liu, Zhenda Xie, and Yebin Liu.
\newblock Dreamcraft3d: Hierarchical 3d generation with bootstrapped diffusion prior.
\newblock \emph{arXiv preprint arXiv:2310.16818}, 2023.

\bibitem[Tang et~al.(2023)Tang, Ren, Zhou, Liu, and Zeng]{tang2023dreamgaussian}
Jiaxiang Tang, Jiawei Ren, Hang Zhou, Ziwei Liu, and Gang Zeng.
\newblock Dreamgaussian: Generative gaussian splatting for efficient 3d content creation.
\newblock \emph{arXiv preprint arXiv:2309.16653}, 2023.

\bibitem[Wang et~al.(2023{\natexlab{a}})Wang, Du, Li, Yeh, and Shakhnarovich]{wang2023score}
Haochen Wang, Xiaodan Du, Jiahao Li, Raymond~A Yeh, and Greg Shakhnarovich.
\newblock Score jacobian chaining: Lifting pretrained 2d diffusion models for 3d generation.
\newblock In \emph{Proceedings of the IEEE/CVF Conference on Computer Vision and Pattern Recognition}, pages 12619--12629, 2023{\natexlab{a}}.

\bibitem[Wang et~al.(2021)Wang, Liu, Liu, Theobalt, Komura, and Wang]{wang2021neus}
Peng Wang, Lingjie Liu, Yuan Liu, Christian Theobalt, Taku Komura, and Wenping Wang.
\newblock Neus: Learning neural implicit surfaces by volume rendering for multi-view reconstruction.
\newblock \emph{arXiv preprint arXiv:2106.10689}, 2021.

\bibitem[Wang et~al.(2023{\natexlab{b}})Wang, Zhang, Zhang, Gu, Bao, Baltrusaitis, Shen, Chen, Wen, Chen, et~al.]{wang2023rodin}
Tengfei Wang, Bo Zhang, Ting Zhang, Shuyang Gu, Jianmin Bao, Tadas Baltrusaitis, Jingjing Shen, Dong Chen, Fang Wen, Qifeng Chen, et~al.
\newblock Rodin: A generative model for sculpting 3d digital avatars using diffusion.
\newblock In \emph{Proceedings of the IEEE/CVF Conference on Computer Vision and Pattern Recognition}, pages 4563--4573, 2023{\natexlab{b}}.

\bibitem[Wang et~al.(2023{\natexlab{c}})Wang, Qian, Huo, Huang, Zhao, and Fu]{yu2023pushing}
Yu Wang, Xuelin Qian, Jingyang Huo, Tiejun Huang, Bo Zhao, and Yanwei Fu.
\newblock Pushing the limits of 3d shape generation at scale, 2023{\natexlab{c}}.

\bibitem[Wang et~al.(2023{\natexlab{d}})Wang, Lu, Wang, Bao, Li, Su, and Zhu]{wang2023prolificdreamer}
Zhengyi Wang, Cheng Lu, Yikai Wang, Fan Bao, Chongxuan Li, Hang Su, and Jun Zhu.
\newblock Prolificdreamer: High-fidelity and diverse text-to-3d generation with variational score distillation.
\newblock \emph{arXiv preprint arXiv:2305.16213}, 2023{\natexlab{d}}.

\bibitem[Wikipedia(2023)]{wiki:janus}
Wikipedia.
\newblock Janus --- wikipedia{,} the free encyclopedia, 2023.
\newblock [Online; accessed 17-November-2023].

\bibitem[Xu et~al.(2023)Xu, Jiang, Wang, Fan, Wang, and Wang]{xu2023neurallift}
Dejia Xu, Yifan Jiang, Peihao Wang, Zhiwen Fan, Yi Wang, and Zhangyang Wang.
\newblock Neurallift-360: Lifting an in-the-wild 2d photo to a 3d object with 360deg views.
\newblock In \emph{Proceedings of the IEEE/CVF Conference on Computer Vision and Pattern Recognition}, pages 4479--4489, 2023.

\bibitem[Yang et~al.(2023)Yang, Cheng, Duan, Ji, and Li]{yang2023consistnet}
Jiayu Yang, Ziang Cheng, Yunfei Duan, Pan Ji, and Hongdong Li.
\newblock Consistnet: Enforcing 3d consistency for multi-view images diffusion.
\newblock \emph{arXiv preprint arXiv:2310.10343}, 2023.

\bibitem[Yao et~al.(2018)Yao, Luo, Li, Fang, and Quan]{yao2018mvsnet}
Yao Yao, Zixin Luo, Shiwei Li, Tian Fang, and Long Quan.
\newblock Mvsnet: Depth inference for unstructured multi-view stereo.
\newblock In \emph{Proceedings of the European conference on computer vision (ECCV)}, pages 767--783, 2018.

\bibitem[Ye et~al.(2023)Ye, Wang, Li, Shi, and Wang]{ye2023consistent}
Jianglong Ye, Peng Wang, Kejie Li, Yichun Shi, and Heng Wang.
\newblock Consistent-1-to-3: Consistent image to 3d view synthesis via geometry-aware diffusion models.
\newblock \emph{arXiv preprint arXiv:2310.03020}, 2023.

\bibitem[Zhang et~al.(2022)Zhang, Bi, Sunkavalli, Su, and Xu]{zhang2022nerfusion}
Xiaoshuai Zhang, Sai Bi, Kalyan Sunkavalli, Hao Su, and Zexiang Xu.
\newblock Nerfusion: Fusing radiance fields for large-scale scene reconstruction.
\newblock In \emph{Proceedings of the IEEE/CVF Conference on Computer Vision and Pattern Recognition}, pages 5449--5458, 2022.

\bibitem[Zhou et~al.(2023)Zhou, Wang, Ma, Liu, Huang, and Wang]{uni3d}
Junsheng Zhou, Jinsheng Wang, Baorui Ma, Yu-Shen Liu, Tiejun Huang, and Xinlong Wang.
\newblock Uni3d: Exploring unified 3d representation at scale.
\newblock \emph{arXiv preprint arXiv:2310.06773}, 2023.

\end{thebibliography}

\end{document}